\RequirePackage{fix-cm}

\documentclass[twocolumn]{svjour3}          % twocolumn
\smartqed  % flush right qed marks, e.g. at end of proof
\usepackage{graphicx}
\usepackage{tabularx}
\usepackage{natbib}
\usepackage{multirow}
\usepackage{amsmath}
\usepackage{bbm}
\usepackage{bm}
\usepackage{times}
\usepackage{booktabs}
\usepackage{array} % for extended column definitions, e.g. >{\em}c
\usepackage{blindtext}
\usepackage{comment}
\usepackage[breaklinks=true]{hyperref}
\usepackage{caption, subcaption}
\usepackage{stfloats}
\usepackage{makecell}
\usepackage{algorithmic}
\usepackage{algorithm}
\usepackage{textcomp}
\usepackage{url}
\usepackage{xcolor}
\usepackage{colortbl}
\usepackage{enumitem,amssymb}
\usepackage{amsfonts}       % blackboard math symbols
\hypersetup{
    colorlinks=true,
    linkcolor=blue,
    filecolor=blue,      
    urlcolor=blue,
    citecolor=blue,
}
\usepackage{pifont}
\usepackage[dvipsnames]{xcolor}

\definecolor{citecolor}{HTML}{0071bc}
\definecolor{tabhighlight}{HTML}{e5e5e5}
\makeatletter
\renewcommand\paragraph{
  \@startsection{paragraph} % name
  {4} % level
  {\z@} % indent
  {.5em \@plus1ex \@minus.2ex} % beforeskip
  {-.5em} % afterskip
  {\normalfont\normalsize\bfseries} % style
}
\makeatother
\begin{document}
\sloppy

\title{Neural Discrimination-Prompted Transformers for Efficient UHD Image Restoration and Enhancement
}

\author{Cong Wang$^{1,2}$ \and
        Jinshan Pan$^{3}$ \and
        Liyan Wang$^{4}$ \and
        Wei Wang$^{1}$ 
        \and
        Yang Yang$^{5}$ 
}

\institute{Cong Wang \at
              \email{supercong94@gmail.com}
           \and
           Jinshan Pan \at
              \email{sdluran@gmail.com}
            \and
            Liyan Wang \at
              \email{wangliyan@mail.dlut.edu.cn}
           \and
          Wei Wang \at
              \email{wangwei29@mail.sysu.edu.cn}
            \and
          Yang Yang \at
              \email{yang.yang4@ucsf.edu}
           \\
           \\
           $^1$ Shenzhen Campus of Sun Yat-Sen University, China  
           \\
           $^2$ The Hong Kong Polytechnic University, Hong Kong  
           \\
           $^3$ Nanjing University of Science and Technology, China 
           \\
           $^4$ Dalian University of Technology, China
           \\
           $^4$ University of California, San Francisco, USA
}
\date{Received: date / Accepted: date}
\maketitle
\begin{abstract} \label{abstract}
 We propose a simple yet effective UHDPromer, a neural discrimination-prompted Transformer, for Ultra-High-Definition (UHD) image restoration and enhancement. Our UHDPromer is inspired by an interesting observation that there implicitly exist neural differences between high-resolution and low-resolution features, and exploring such differences can facilitate low-resolution feature representation. To this end, we first introduce Neural Discrimination Priors (NDP) to measure the differences and then integrate NDP into the proposed Neural Discrimination-Prompted Attention (NDPA) and Neural Discrimination-Prompted Network (NDPN). The proposed NDPA re-formulates the attention by incorporating NDP to globally perceive useful discrimination information, while the NDPN explores a continuous gating mechanism guided by NDP to selectively permit the passage of beneficial content. To enhance the quality of restored images, we propose a super-resolution-guided reconstruction approach, which is guided by super-resolving low-resolution features to facilitate final UHD image restoration. Experiments show that UHDPromer achieves the best computational efficiency while still maintaining state-of-the-art performance on $3$ UHD image restoration and enhancement tasks, including low-light image enhancement, image dehazing, and image deblurring. The source codes and pre-trained models will be made available at \url{https://github.com/supersupercong/uhdpromer}.


\end{abstract}

\section{Introduction}\label{sec:introduction}
The advancement of imaging sensors and displays has notably propelled Ultra-High-Definition (UHD) imaging in recent years. 
However, UHD images which generally mean $3,840 $$\times$$ 2,160$ resolution or 4K images, when captured under challenging conditions such as low light, haze, or rapid motion, often exhibit degradation, compromising visual quality and hindering the performance of high-level vision tasks.
This paper addresses these challenges by introducing a general framework to recover degraded UHD images.

The advent of convolutional neural networks (CNNs)~\citep{AlexNet,res,he2016identity,xie2017aggregated,dense_network_cvpr17,liu2022convnet_convnext,woo2023convnext} and Transformers~\citep{vision_transformer,liu2021swin,Liu_2022_CVPR,Liu_2022_CVPR_Video_Swin_Transformer} has heralded significant achievements in image restoration~\citep{dehazing_tip16_dehazenet,DnCNN,RCAN,dehaze_cvpr18_gan,Zamir_2021_CVPR_mprnet,physicsgan_pan,pan_pami_l0,pan2023cascaded,ren2021deblurring,bai2024self,wang_aaai22,Tsai2022Stripformer,pan2022dual_ijcv,promptrestorer,SelfPromer,msgnn}. 
Despite these advancements, existing methods primarily focus on general image restoration tasks, yet they struggle with processing UHD image sizes due to computational constraints and effectively improving UHD image quality.
Such limitations restrict the potential applications of these algorithms in UHD imaging systems.

Recently, there has been a surge in UHD restoration approaches to meet the demands associated with handling degraded UHD images. 
Notable methods, such as~\cite{Zheng_uhd_CVPR21,Li2023ICLR_uhdfour}, employ intricate deep neural networks that primarily focus on learning from low-resolution features by down-sampling high-resolution counterparts, enabling the management of larger UHD image sizes. 
However, these methodologies often overlook the critical exploration of neural differences existing between high-resolution and low-resolution features, which implicitly contain valuable information to guide low-resolution learning for better UHD image restoration. 
Neglecting this aspect may substantially diminish the quality of the reconstruction performance.
Recent work by \cite{uhdformer} demonstrates that high-resolution features contain more useful information that can effectively guide low-resolution feature learning. 
Building upon this insight, they propose UHDformer, which facilitates knowledge transfer from high-resolution features to low-resolution ones through correlation matching-based feature transformation.
Although this approach achieves excellent performance with a very light-weight model, this approach exhibits two major limitations.
First, the correlation computation relies on matrix similarity operations in high-dimensional feature space, which inevitably introduces increased computational overhead and compromises model efficiency. 
Hence, a more efficient model, which can keep superior performance, is needed.
Second, UHDformer focuses exclusively on direct feature transformation while neglecting to explore the in-depth relationships and hierarchical dependencies between high-resolution features and low-resolution ones.
Hence, building a more interpretable and insightful model is still worth exploring.

To solve the above challenges, we propose a neural discrimination-prompted Transformer, termed UHDPromer, for more efficient UHD image restoration and enhancement. 
Our UHDPromer is inspired by an interesting observation that there implicitly exist neural differences between high-resolution features and low-resolution ones, which could provide the low-resolution domain with useful information to improve low-resolution feature representation. 
Specifically, we introduce the Neural Discrimination Priors (NDP) to measure the neural differences.
Based on the NDP, we develop a Neural Discrimination-Prompted Transformer (NDPT), which integrates NDP into Transformer designs to effectively learn more representative low-resolution features.
The NDPT comprises two key components: Neural Discrimination-Prompted Attention (NDPA) and Neural Discrimination-Prompted Network (NDPN).
The NDPA reconsiders the cross-attention between NDP and the query vector derived from low-resolution features, which is further formed into new attention with rest key and value vectors obtained from low-resolution features to effectively perceive more useful discrimination information within the perspective of long-range pixel dependencies. 
The NDPN explores a continuous gating mechanism, leveraging the NDP to consecutively guide the feed-forward encoding process. 
It enables more beneficial content to be transmitted within a pixel-wise gating modulation manner, thereby enhancing the low-resolution feature representation ability.

One natural way to reconstruct final clean images is to fuse deep features from the low-resolution branch and the deep features from the high-resolution branch.
However, we note that such a manner cannot ensure good results.
To solve this problem, we further propose a super-resolution-guided reconstruction method, where we initially super-resolve low-resolution features learned in the NDPT, and then utilize these super-resolution features to guide the final UHD image reconstruction.
With the above designs, our UHDPromer achieves state-of-the-art performance on various UHD image restoration and enhancement tasks while exhibiting efficiency.

We summarize our main contributions as follows:
\begin{itemize}
\item We propose a simple, effective, and strong UHDPromer by exploring the neural discrimination priors implicitly implied between high-resolution features and low-resolution ones for efficient UHD image restoration.
\item We propose a fruitful neural discrimination-prompted attention and a valid neural discrimination-prompted network, which integrates the neural discrimination priors into attention and feed-forward network designs.
\item Experiments show that our UHDPromer is computationally efficient while still generating the best restoration quality under different training settings on 3 UHD image restoration and enhancement tasks, including low-light enhancement, dehazing, and deblurring.
\end{itemize}

\section{Related Work}
In this section, we review the most relevant image restoration approaches including CNN-based and Transformers-based techniques.
We also discuss the existing methods for UHD image restoration and enhancement. 
\textbf{\subsection{CNN-based Image Restoration Approaches}}
CNN-based architectures~\citep{grid_dehaze_liu,zhao_lie,dualcnn_cvpr18,derain_jorder_yang,fu2017removing,dehazing_tip16_dehazenet,dehazing_eccv16_mscnn,dcsfn-wang-mm20,cho2021rethinking_mimo,Zamir_2021_CVPR_mprnet,msbdn_cvpr20_dong,dmphn2019,anwar2019deep,li2019single,tian2020deep} have been demonstrated to surpass conventional restoration approaches~\citep{dehazing_tpami11_darkchannel,dehazing_TOG14_colorline,pan2016blind,pan_pami_l0,dehazing_cvpr16_nonlocal,derain_lp_li,Gamma_tip,derain_lowrank,derain_dsc_luo} by implicitly learning priors from large-scale data.
Most CNN architectures are designed based on residual learning, dense connections, the encoder-decoder, and attention-based models.
For example,
\cite{cho2021rethinking_mimo} present a multi-input multi-output MIMO-UNet by encoding multi-scale input images and decoding multiple deblurred images with different scales to merge multi-scale features efficiently. 
To remedy the missing spatial information from high-resolution features,~\cite{msbdn_cvpr20_dong} propose a multi-scale boosted dehazing network with a dense feature fusion module based on the U-Net.
residual learning with skip connections has been used for image deblurring~\citep{deblurgan_cvpr18}, and dense connections for image dehazing~\citep{msbdn_cvpr20_dong}.
The encoder-decoder~\citep{cho2021rethinking_mimo}, and attention-based~\citep{grid_dehaze_liu} models are often used to extract multi-scale information. \cite{grid_dehaze_liu} propose an end-to-end trainable CNN, named GridDehazeNet, for single image dehazing, which implements an attention-based multi-scale estimation to extract multi-scale information. 
Subsequently,~\cite{Zamir_2021_CVPR_mprnet} propose a multi-stage image restoration method MPRNet, which achieved great progress by progressively recovering clean images by learning a subnetwork by supervised attention at each stage.
Although CNN-based methods have achieved excellent results, they still lack the ability to model global information due to mainly learning local features.
\textbf{\subsection{Transformers-based Image Restoration 
 Methods}}
Transformer-based models~\citep{liang2021swinir,wang2021uformer,Zamir2021Restormer,Tsai2022Stripformer,Kong_2023_CVPR_fftformer,DehazeFormer,chen2021IPT} have achieved decent results in image restoration. 
They excel by modeling long-range pixel dependencies, overcoming the limitations of CNN-based algorithms that perform computations in local windows.
~\cite{liang2021swinir} propose a strong baseline model named SwinIR for image restoration based on the Swin Transformer. 
~\cite{Zamir2021Restormer} also propose an efficient Transformer model named Restormer using a U-net structure including multi-head attention and feed-forward network, and achieved state-of-the-art results on several image restoration tasks.
In addition,~\cite{wang2021uformer} present a hierarchical encoder-decoder network Uformer using the Transformer block for image restoration by adopting a locally-enhanced window (LeWin) Transformer block and learnable multi-scale restoration modulator.
However, despite their promising performance in general image restoration tasks, these Transformer-based methods typically struggle with UHD images due to relying on many parameters, heavy computation, and high MACs, constraining their broader applicability to UHD imaging devices.
\textbf{\subsection{UHD Image Restoration and Enhancement}}
% {\flushleft \bf{UHD image restoration and enhancement.}}
The burgeoning demand for processing UHD images in imaging systems has spurred the development of a myriad of innovative methods to recover clear UHD images. 
A plethora of network designs have emerged in response, such as bilateral learning for image dehazing~\citep{Zheng_uhd_CVPR21}, multi-scale separable-patch integration networks for video deblurring~\citep{uhd_video_deblurring}, and Fourier embedding networks for low-light image enhancement~\citep{Li2023ICLR_uhdfour}.
Drawing inspiration from the bilateral learning approach, ~\cite{Zheng_uhd_CVPR21} innovatively propose multi-guided bilateral learning dedicated to the dehazing of UHD images, which uses deep CNN to build an affine bilateral grid to maintain detailed edges and textures in the image.
In a parallel vein,~\cite{uhd_video_deblurring} forge ahead by crafting multi-scale networks infused with a separable-patch integration scheme, a strategic move to attain an expansive receptive field crucial for UHD deblurring.
Venturing into the realm of low-light image enhancement,~\cite{LLformer} unveil a Transformer architecture endowed with axis-based multi-head self-attention and cross-layer attention fusion to reduce the linear complexity.
Driven by compelling peculiarities unveiled in the Fourier domain,~\cite{Li2023ICLR_uhdfour} champion the integration of the Fourier transform into a cascading network, adeptly achieving amplitude and phase enhancement by leveraging unique implementation strategies. 
\cite{uhdformer} propose the first Transformer architecture, UHDformer, for the UHD image restoration problem.
The authors find that the high-resolution features can provide more useful information to improve the representation of low-resolution features.
Based on these findings, the authors propose to build the transformation from high-resolution features to low-resolution ones by proposing the correlation matching between them.
In \citep{yu2024empowering}, \citeauthor{yu2024empowering} design a customized resampling under the guidance of model knowledge, integrating the resizers and enhancers through compensatory information learning to solve the UHD image enhancement problem. 
Instead of focusing solely on the input image’s intrinsic features,~\cite{uhddip} propose a dual interaction prior-driven UHD restoration network that integrates the additional normal prior and gradient prior in the low-resolution space for dynamically enhancing the structures and details to guide high-resolution feature restoration.

In this paper, we conduct an in-depth investigation into neural discrimination priors by analyzing the feature disparities between high-resolution and low-resolution representations. 
Our key insight is that these disparities encode critical structural information that can be leveraged to improve restoration quality. 
Based on this finding, we propose a neural discrimination-prompted Transformer architecture specifically designed for UHD image restoration and enhancement.
The proposed method achieves superior restoration performance compared to existing approaches while maintaining computational efficiency. 
Extensive experimental validation demonstrates the effectiveness of incorporating neural discrimination priors into the Transformer framework for UHD image processing tasks.

\section{Proposed Approach}

Our method is inspired by the existence of neural differences between high-resolution and low-resolution features.
Such differences implicitly contain useful structures, which can serve as a prompt to guide the learning of low-resolution features.
Hence, we introduce the  Neural Discrimination Priors (NDP) to measure the differences.
Furthermore, we develop a new Neural Discrimination-Prompted Transformer, which integrates NDP into Transformer designs to effectively learn low-resolution features.
Moreover, we propose a super-resolution-guided reconstruction method to guide the final UHD image restoration.
\begin{figure*}[!t]
% \scriptsize
%\vspace{-5mm}
\centering
\begin{center}
\begin{tabular}{c}
\includegraphics[width=1\linewidth]{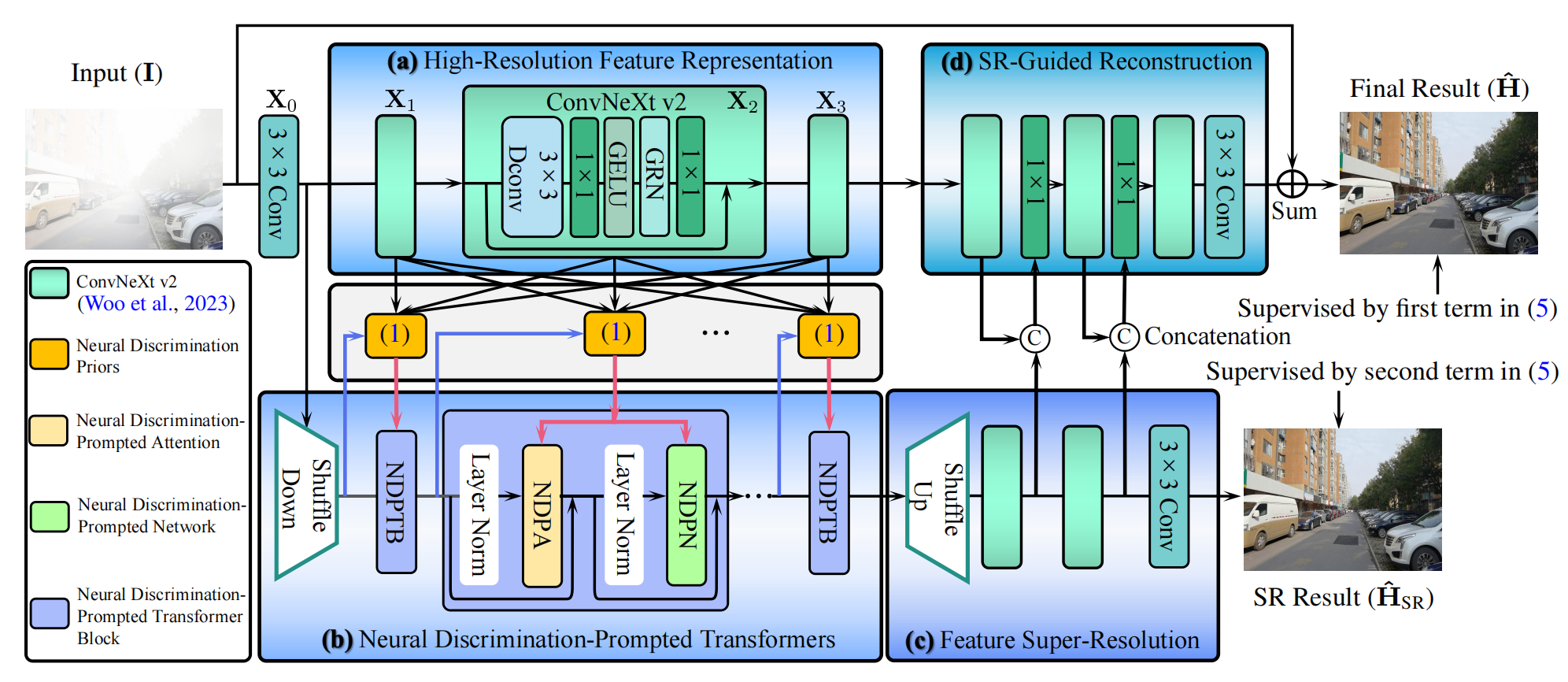} 
\end{tabular}

\caption{\textbf{Overview of our proposed Neural Discrimination-Prompted Transformers (UHDPromer)}.
Our UHDPromer contains $4$ parts: \textbf{(a)} High-Resolution Feature Representation (HRFR); \textbf{(b)} Neural Discrimination-Prompted Transformers (NDPT); \textbf{(c)} Feature Super-Resolution (FeaSR); and \textbf{(d)} SR-Guided Reconstruction (SRG-Recon).
Firstly, HRFR explores hierarchical multi-scale high-resolution features, which participate in forming neural discrimination priors (NDP) with low-resolution features. 
Then, NDPT, which is guided by NDP, learns low-resolution features.
Next, FeaSR super-resolves the output features of NDPT.
Finally, SRG-Recon reconstructs final images guided by learned SR features in FeaSR.
}
\label{fig: Framework of UHDPromer}
\end{center}
% \vspace{-3mm}
\end{figure*}
\textbf{\subsection{Overall Pipeline}}\label{sec:Overall Pipeline}
Fig.~\ref{fig: Framework of UHDPromer} shows the overall framework of our UHDPromer, which consists of four parts: \textbf{(a)} High-Resolution Feature Representation (HRFR); \textbf{(b)} Neural Discrimination-Prompted Transformers (NDPT); \textbf{(c)} Feature Super-Resolution (FeaSR); and \textbf{(d)} SR-Guided Reconstruction (SRG-Recon).
Given a UHD input image $\mathbf{I}$~$\in$~$\mathbb{R}^{H \times W \times 3}$, we first apply a $3$$\times$$3$ convolution to obtain low-level embeddings $\mathbf{X}_0$~$\in$~$\mathbb{R}^{H \times W \times C}$, where $H\times W$ denotes the spatial dimension and $C$ is the number of channels. 
Then, HRFR hierarchically encodes the shallow features $\mathbf{X}_{0}$ to explore multi-scale features $\{\mathbf{X}_1, \mathbf{X}_2, \mathbf{X}_3\}$~$\in$~$\mathbb{R}^{H \times W \times C}$ via $3$ ConvNeXt-v2 blocks~\citep{woo2023convnext}.
Next, NDPT receives the shuffle down features $\mathbf{X}_{\textrm{down}}$~$\in$~
% $\mathbb{R}^{\frac{H}{s} \times \frac{W}{s} \times C}$ 
$\mathbb{R}^{H/s \times W/s \times C}$
from $\mathbf{X}_{0}$, and $\mathbf{X}_{\textrm{down}}$ is subsequently sent to Neural Discrimination-Prompted Transformers (NDPT) to learn low-resolution features.
Here, $s$ denotes the shuffle down factor.
At the same time, NDPT is also guided by the developed Neural Discrimination Priors (NDP), which can measure useful discrimination content between high-resolution features and low-resolution ones to learn more representative features for better UHD image restoration.
After that, FeaSR super-resolves the output features of NDPT and generates super-resolution images $\mathbf{\hat{H}}_{\textrm{SR}}$~$\in$~$\mathbb{R}^{H\times W \times 3}$.
Finally, SRG-Recon, which contains three ConvNeXt-v2 blocks~\citep{woo2023convnext}, two 1$\times$1 convolutions, and a 3$\times$3 convolution, receives the output features from HRFR and super-resolution features from FeaSR and generates residual image $\mathbf{S}$~$\in$~$\mathbb{R}^{H\times W \times 3}$ to which UHD input image is added to obtain the restored image: $\mathbf{\hat{H}} = \mathbf{I} + \mathbf{S}$.
\begin{figure*}[!t]
% \vspace{-2mm}
% \scriptsize
%\vspace{-5mm}
\centering
\begin{center}
\begin{tabular}{ccccccccc}
\includegraphics[width=1\linewidth]{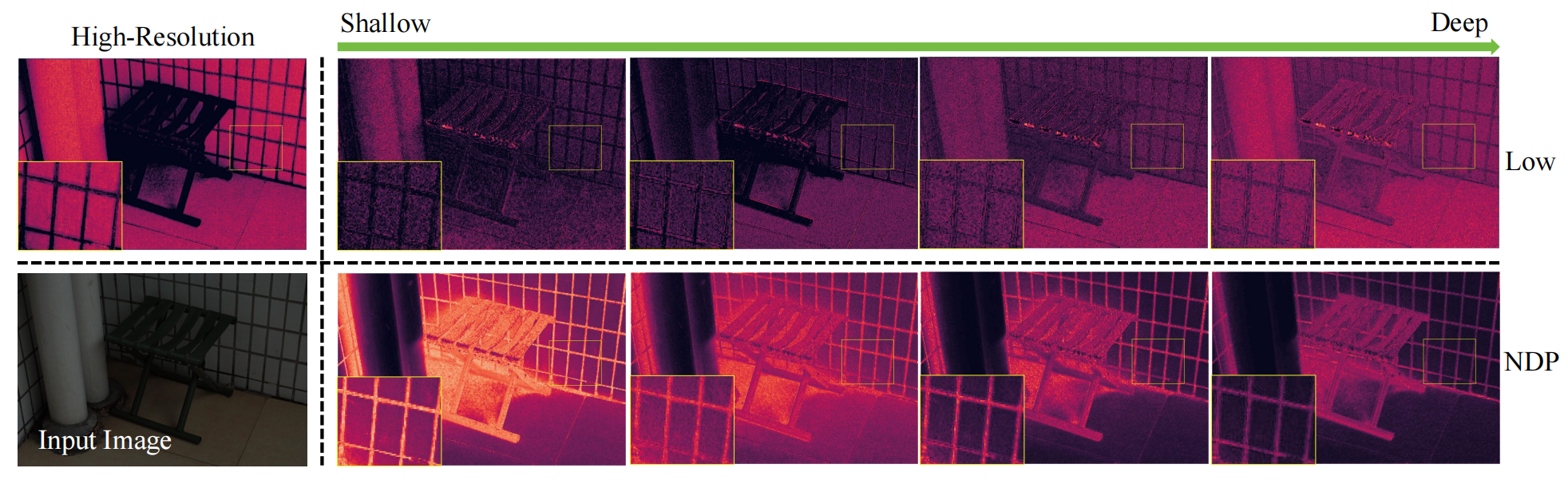} 
\end{tabular}
% \vspace{-2mm}
\caption{\textbf{Illustration of the Neural Discrimination Priors} (NDP) in the different layers by \eqref{eq: NDP}.
We note that the low-resolution features exist obvious content differences compared with high-resolution ones.
Specifically, low-resolution features always hand down extensive noises while some structures are blurry.
Our NDP can better identify these regions, which can effectively measure whether the structures lie in shallower layers or deeper layers, serving as the discrimination function to help learn more representative features in the low-resolution domain.
}
\label{fig:Illustration of Implicit Neural Discrimination Priors}
\end{center}
% \vspace{-5mm}
\end{figure*}
\textbf{\subsection{Neural Discrimination Prior}}\label{sec:Exact Correlation Matching Transformer Block}
As the low-resolution features are derived by shuffling down the shallow high-resolution embeddings, it would inevitably raise some structural differences compared with the input images.
Hence, how to correct the differences and utilize them to better guide the learning of low-resolution features presents a challenging and worthwhile area for exploration.
In this paper, we introduce the Neural Discrimination Priors implicitly implied between high-resolution features and low-resolution ones, which would measure the differences mentioned above. 
Integrating the priors into the learning of low-resolution domains can improve the low-resolution feature representation ability. 
The Neural Discrimination Prior (NDP) is defined as:
%
%\vspace{-2mm}
% \\
\begin{equation}
\begin{split}
\textit{NDP}_{i}(x) = 1/\sqrt{e^{\textrm{abs}\left| \mathcal{H}_{i}\left[\mathbf{X}_1, \mathbf{X}_2, \mathbf{X}_3\right]\left(x\right)- \mathbf{Y}_{i}\left(x\right) \right|}}
\label{eq: NDP}
\end{split}
% \vspace{-0.95em}
\end{equation}
where $\mathcal{H}_{i}[\cdot,\cdot,\cdot]$ is composed of a concatenation operation and a stride convolution with $s$$\times$$s$ kernel sizes and $s$$\times$$s$ stride sizes, while $x$ is the pixel position;
$\textrm{abs}|\cdot|$ means the absolute value operation.
$\mathbf{Y}_{i}$ denotes the low-resolution input feature of $i$-th Transformer block, where $i = 1, 2, \dots, L$.
Equation \eqref{eq: NDP} suggests that when the value of $\textit{NDP}_{i}(x)$ approaches $1$, the feature at position $x$ notably diverges from low-resolution features, indicating greater discriminative potential.
Hence, we can use $\textit{NDP}_{i}(x)$ to help Transformers better distinguish whether the feature is discriminative so that Transformers can pay more attention to useful content in low-resolution space for better restoration.

Fig.~\ref{fig:Illustration of Implicit Neural Discrimination Priors} shows the visual features map of the high-resolution feature, low-resolution feature, and NDP.
It is obviously evident that the NDP can better describe much clearer content than the low-resolution one, which thus facilitates guiding deep models to better learn low-resolution features for better restoration.
\begin{figure*}[t]
% \scriptsize
%\vspace{-5mm}
\centering
\begin{center}
\begin{tabular}{c}
\hspace{-1.5mm}\includegraphics[width=1\linewidth]{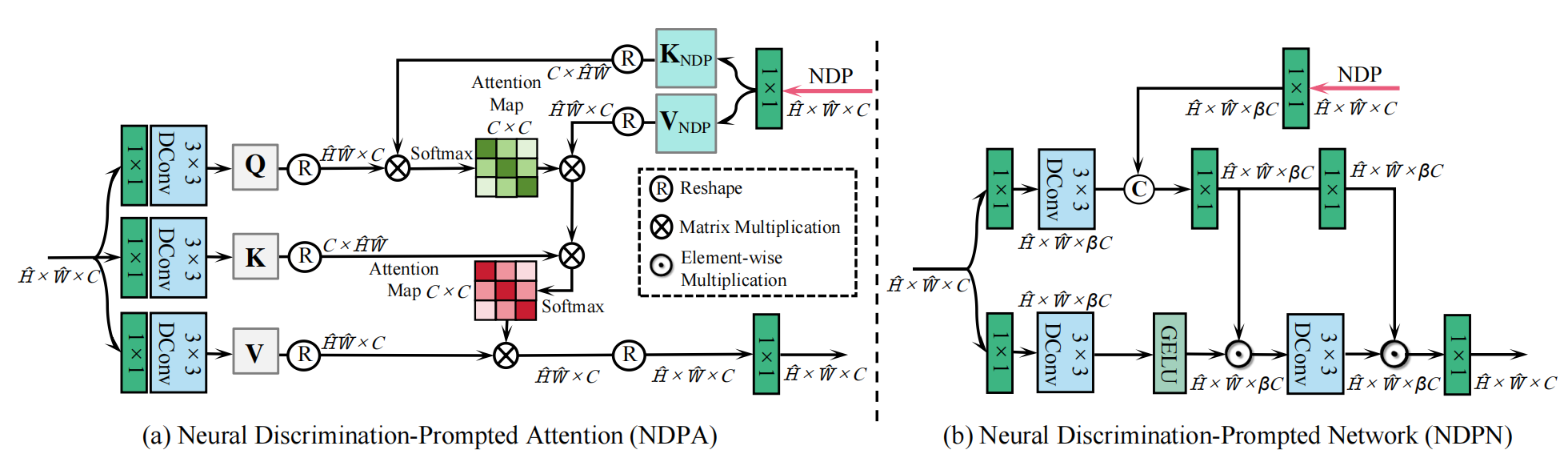} 
\end{tabular}
% \vspace{-2mm}
\caption{\textbf{(a)} Neural Discrimination-Prompted Attention (\textbf{NDPA}) and \textbf{(b)} Neural Discrimination-Prompted Network (\textbf{NDPN}).
The NDPA first performs cross attention between NDP and \emph{query} vector generated by low-resolution features to globally perceive the NDP, which would be re-performed by the attention computation between rest \emph{key} and \emph{value} vectors of low-resolution features to further explore useful learned information.
% %
The NDPN considers the continuous gating mechanism guided by the NDP to allow useful information to be passed to facilitate final image reconstruction.
}
\label{fig: (a) Neural Discrimination-Prompted Attention (NDPA) and (b) Neural Discrimination-Prompted Network (NDPN)}
\end{center}
% \vspace{-5mm}
\end{figure*}
\textbf{\subsection{Neural Discrimination-Prompted Transformer}}\label{sec: Neural Discrimination-Prompted Transformer}
To better learn low-resolution features, we propose a sample and effective Neural Discrimination-Prompted Transformer Block (NDPTB) that is guided by the developed NDP, containing Neural Discrimination-Prompted Attention (NDPA) and Neural Discrimination-Prompted Network (NDPN).
Our Neural Discrimination-Prompted Attention (NDPA) and Neural Discrimination-Prompted Network (NDPN) modules are inspired by the multi-Dconv head transposed attention (MDTA) and gated-Dconv feed-forward network (GDFN) architectures from Restormer~\citep{Zamir2021Restormer}.
However, our approach introduces fundamental architectural innovations. 
Unlike Restormer's conventional attention mechanism, NDPA dynamically incorporates neural discrimination priors (NDP) into the attention computation through continuous reformation and reorganization of the query-key-value (QKV) representations. 
This allows the attention mechanism to adaptively focus on discriminative features that are crucial for high-quality reconstruction. 
Similarly, NDPN extends the integration of NDP into the feed-forward network by implementing a continuous gating mechanism guided by discrimination priors. 
This design enables the feed-forward network to selectively process features based on their discriminative importance, thereby enhancing the overall reconstruction quality.
Each NDPTB not only receives the output of the previous NDPTB but also the NDP ($\textrm{$\textbf{Y}$}_{\text{NDP}}$) to guide the low-resolution learning:
\begin{equation}
% \footnotesize
\label{eq: NDPTB}
\begin{split}
&\textrm{$\textbf{X}$}^{'} = \textit{NDPA}\Big(\textit{LN}(\textrm{$\textbf{X}$}^{i}), \textrm{$\textbf{Y}$}_{\text{NDP}} \Big) + \textrm{$\textbf{X}$}^{i},
\\
&\textrm{$\textbf{X}$}^{i+1} = \textit{NDPN}\Big(\textit{LN}(\textrm{$\textbf{X}$}^{'}), \textrm{$\textbf{Y}$}_{\text{NDP}}\Big) + \textrm{$\textbf{X}$}^{'}, % \label{eq:ecm-tb2}
\end{split}
\end{equation}
% \\
%\vspace{-2mm}
where $\textrm{$\textbf{X}$}^{i}$ means the output features of $i^{\text{th}}$ NDPTB;
$\textit{LN}(\cdot)$ means the layer normalization~\citep{ba2016layer}.
\\
\subsubsection{Neural Discrimination-Prompted Attention}

NPDA, as shown in Fig.~\ref{fig: (a) Neural Discrimination-Prompted Attention (NDPA) and (b) Neural Discrimination-Prompted Network (NDPN)}(a), effectively integrates the NDP into attention to perceive the useful discrimination information within global perspectives.
Unlike existing attention designs~\citep{Zamir2021Restormer,liu2021swin}, which only compute single attention, we propose a continuous attention computation scheme to adequately utilize the NDP by long-range pixel dependency modeling.
Specifically, NPDA first computes the cross-attention between the NDP features $\textrm{$\textbf{Y}_{\textrm{NDP}}$}$ and the \emph{query} vector $\textbf{Q}$ generated by low-resolution splitting features $\textrm{$\hat{\textbf{X}}$}$.
Subsequently, the attention map undergoes a re-computation of attention, utilizing the remaining \emph{key} vector $\textbf{K}$ and \emph{value} vector $\textbf{V}$ from the low-resolution split features. This process leverages the aggregated discriminatory information in cross-attention to effectively guide low-resolution learning.
The NPDA can be subsequently computed as follows:
\begin{equation}
% \footnotesize
\label{eq: NDPA}
\begin{split}
&\textbf{Q}, \textbf{K}, \textbf{V} = \mathcal{S}\Big(W_{d}W_{p}(\textrm{$\hat{\textbf{X}}$})\Big),
\\
&\textbf{K}_{\textrm{NDP}}, \textbf{V}_{\textrm{NDP}} = \mathcal{S}\Big(W_{d}W_{p}(\textrm{$\textbf{Y}_{\textrm{NDP}}$})\Big),
\\
% &\textrm{$\textbf{X}_{\textrm{NPDA}}$} =  \mathcal{A}\bigg
% (\mathcal{A}\Big(\textbf{Q}, \textbf{K}_{\textrm{NDP}}, \textbf{V}_{\textrm{NDP}}\Big), \textbf{K}, \textbf{V}
% \bigg),
&\textrm{$\textbf{X}_{\textrm{NPDA}}$} =  \mathcal{A}\bigg
(\mathcal{A}\Big(\textbf{Q}, \textbf{K}_{\textrm{NDP}}, \textbf{V}_{\textrm{NDP}}\Big), \textbf{K}, \textbf{V}
\bigg),
\end{split}
% \vspace{-0.5em}
\end{equation}
% \\
% \vspace{-8mm}
% \\
where $\mathcal{A}\left(\hat{\textbf{Q}}, \hat{\textbf{K}}, \hat{\textbf{V}}\right) = \hat{\textbf{V}}\cdot \textrm{Softmax$\left( \hat{\textbf{K}} \cdot \hat{\textbf{Q}}/\alpha \right)$}$; 
Here, $\alpha$ is a learnable scaling parameter to control the magnitude of the dot product of $\hat{\mathbf{K}}$ and $\hat{\mathbf{Q}}$;
$\mathcal{S}(\cdot)$ denotes the split operation;
$W_{p}(\cdot)$ refers to the $1$$\times$$1$ point-wise convolution, while $W_{d}(\cdot)$ means the $3$$\times$$3$ depth-wise convolution;
$\textrm{$\textbf{X}_{\textrm{NPDA}}$}$ means the output of the NPDA.

\subsubsection{Neural Discrimination-Prompted Network}
We propose the NDPN, as shown in Fig.~\ref{fig: (a) Neural Discrimination-Prompted Attention (NDPA) and (b) Neural Discrimination-Prompted Network (NDPN)}(b), to explore the continuous gating mechanism guided by the NDP to allow more beneficial information to be passed.
Different from existing feed-forward networks~\citep{Zamir2021Restormer,liang2021swinir,chen2021IPT,wang2021uformer} that only consider a single gating scheme or point-wise perception, we suggest a continuous gating mechanism, which not only adequately integrates the NDP into feed-forward network designs within a pixel-level modulation manner but also explores the high-order gating to enable more discriminative information and better guidance of low-resolution learning.
Specifically, we first use $1$$\times$$1$ convolution, and $3$$\times$$3$ depth-wise convolution to process input low-resolution features, which would be split into two parts: $\textrm{$\textbf{Z}_{1}$}$ and $\textrm{$\textbf{Z}_{2}$}$.
Then, we fuse the NDP (\textrm{$\textbf{Y}_{\textrm{NDP}}$}) with $\textrm{$\textbf{Z}_{1}$}$ by concatenation operation and $1$$\times$$1$ convolution to generate the fused features $\textrm{$\textbf{X}_{\textrm{fusion}}$}$.
Next, the GELU is performed on $\textrm{$\textbf{Z}_{2}$}$, which is further gated by $\textrm{$\textbf{X}_{\textrm{fusion}}$}$ to produce the gated feature $\textrm{$\textbf{X}_{\textrm{gate}}^{1}$}$.
After that, the gated feature $\textrm{$\textbf{X}_{\textrm{gate}}^{1}$}$ is input to a $3$$\times$$3$ depth-wise convolution and would be further gated by the fused features processed by $1$$\times$$1$ convolution.
% \xiaoyu{``input" $\rightarrow$ ``inputted"}
%
Last, the gated feature is conducted with $1$$\times$$1$ convolution to produce the final output feature.

The process of NDPN can be subsequently computed by:
% \\
% \vspace{-8mm}
% \\
\begin{equation}
% \footnotesize
\label{eq: NDPN}
\begin{split}
&\textrm{$\textbf{Z}_{1}$}, \textrm{$\textbf{Z}_{2}$} = \mathcal{S}\left(W_{d}W_{p} (\textrm{$\hat{\textbf{X}}$})\right),
\\
&\textrm{$\textbf{X}_{\textrm{fusion}}$} = W_{p}\big(\mathcal{C}\left[\textrm{$\textbf{Z}_{1}$}, \textrm{$\textbf{Y}_{\textrm{NDP}}$}\right]\big),
\\
&\textrm{$\textbf{X}_{\textrm{gate}}^{1}$} = \textrm{$\textbf{X}_{\textrm{fusion}}$} {\odot} \sigma(\textrm{$\textbf{Z}_{2}$}),
\\
&\textrm{$\textbf{X}_{\textrm{NDPN}}$} = W_{p}\Big(W_{d}\left(\textrm{$\textbf{X}_{\textrm{gate}}^{1}$}\right) {\odot} W_{p}\left(\textrm{$\textbf{X}_{\textrm{fusion}}$}\right)\Big),
\end{split}
\end{equation}
% \\
% \vspace{-8mm}
% \\
where $\mathcal{C}[\cdot,\cdot]$ denotes the concatenation operation;
$\sigma(\cdot)$ means the GELU function;
${\odot}$ is the element-wise multiplication;
$\textrm{$\textbf{X}_{\textrm{NDPN}}$}$ means the output of NDPN.
Following \citep{Zamir2021Restormer}, we use an expanding factor $\beta$ to enlarge the channel dimension of intermediate features to scale up model capacity.
\textbf{\subsection{Training Objectives}}\label{sec:Learning Objective}
%\vspace{-2mm}
%
As our UHDPromer involves an SR-guided reconstruction branch and a super-resolution branch, it is necessary to develop loss functions to respectively constrain the training of these two branches for better results. Specifically, let $\mathbf{\hat{H}}$ and $\mathbf{\hat{H}}_{\textrm{SR}}$ denote the outputs of the SR-guided reconstruction branch and the super-resolution branch, $\mathbf{H}$ denotes the ground-truth, we use the following loss function to constrain the network training: 
\begin{equation}
%\footnotesize
% \begin{split}
\label{eq: loss}
%\mathcal{L} = \|I-IIIIII\|_1 + \alpha \|XXxxxxxxxxxx\|, 
\mathcal{L} = \phi(\mathbf{\hat{H}}, \mathbf{H}) + \alpha \phi(\mathbf{\hat{H}}_{\textrm{SR}}, \mathbf{H}), 
% \end{split}
\end{equation}
where $\phi(\cdot,\cdot)$ denotes a spatial- and frequency-domain based loss function used in~\citep{sun2022shufflemixer,cho2021rethinking_mimo,uhdformer}. The weight $\alpha$ is empirically set to be $0.5$.
\\
\section{Experiments}
We evaluate our UHDPromer for $3$ UHD image restoration and enhancement tasks: \textbf{(a)} UHD low-light image enhancement, \textbf{(b)} UHD image dehazing, and \textbf{(c)} UHD image deblurring. 
We also conduct extensive ablation study experiments to analyze the effectiveness as well as discuss the limitations of the proposed UHDPromer.

\textbf{\subsection{Experimental Setup}}
\subsubsection{Implementations}
% \noindent \textbf{Implementations.}
These are 15 NDPTBs, i.e., $L=15$, where the number of attention heads in each NDPTB is $8$.
We set the number of channels $C$ as $16$.
We train models using AdamW optimizer~\citep{adamw} with the initial learning rate $5e^{-4}$ gradually reduced to $1e^{-7}$ with the cosine annealing~\citep{loshchilov2016sgdr}.
The training patch size is set as $512$$\times$$512$.
The shuffle down factor is set as 8, i.e., $s=8$, while both kernel and stride sizes in stride convolution are also set as 8.
\begin{table}[!t]
% \tablestyle{11pt}{1}
\setlength{\tabcolsep}{11.3pt}
\caption{\textbf{Datasets Statistics of UHD image restoration benchmarks}.
}
\label{tab: Datasets Statistics.} 
\begin{tabular}{l|cc|cccc}
\Xhline{1.5pt}
\textbf{Dataset} &\textbf{Training} & \textbf{Testing} & \textbf{Resolution}
\\
\Xhline{1pt}
\textbf{UHD-LL} &2,000 &150 & 3,840 $\times$ 2,160 \\ 
\textbf{UHD-Haze}&2,290 &230 & 3,840 $\times$ 2,160 \\ 
\textbf{UHD-Blur}&1,964 &300 & 3,840 $\times$ 2,160 \\
\Xhline{1.5pt}
\end{tabular}
\end{table}
\begin{table*}[t]
% \scriptsize
%\vspace{-2mm}
\caption{\textbf{UHD low-light image enhancement}. 
Our \textbf{UHDPromer} achieves the state-of-the-art performance under different training settings.
The best and second best are marked in \textbf{bold} and \underline{underlined}, respectively.
}
\label{tab:Low-light image enhancement.} 
%\tablestyle{25.5pt}{1.05}
\setlength{\tabcolsep}{25.5pt}
\begin{tabular}{l|c|cc|l}
\Xhline{1.5pt}
 \textbf{Method} &\textbf{Venue} &   \textbf{PSNR}~$\uparrow$  & \textbf{SSIM}~$\uparrow$  & 
 \textbf{LPIPS}~$\downarrow$
\\
\Xhline{1pt} % \textcolor{blue}{\textit{\textbf{Setting1:}}}
 \multicolumn{5}{c}{\textcolor{blue}{\textit{\textbf{Setting 1:}}} Trained on LOL}\\
\Xhline{1pt}
SwinIR~\citep{liang2021swinir}&ICCVW'21&17.900  &0.7379 & 0.5217 
\\
Restormer~\citep{Zamir2021Restormer}&CVPR'22&19.728 & 0.7703  &0.4566
\\
Uformer~\citep{wang2021uformer}&CVPR'22&18.168 & 0.7201  &0.5593
\\
LLFlow~\citep{wang2021llflow}&AAAI'22&19.596&0.7333& 0.4606

\\
LLformer~\citep{LLformer}&AAAI'23&21.440& 0.7763 &0.4528

\\
UHDFour~\citep{Li2023ICLR_uhdfour}&ICLR'23& 14.771 & 0.3760 &0.7608

\\
LMAR~\citep{yu2024empowering}&CVPR'24&22.430&  \textbf{0.8628} & 0.4211\\
UHDformer~\citep{uhdformer} &AAAI’24&\textbf{22.615}&  0.7754  &\underline{0.4241} \\
\underline{UHDPromer (\textit{Ours})} &-&\underline{21.714}&  \underline{0.7807}  & \textbf{0.4176}
%0.7430M
\\
\Xhline{1pt}
 \multicolumn{5}{c}{\textcolor{blue}{\textit{\textbf{Setting 2:}}} Trained on UHD-LL}\\
\Xhline{1pt}
SwinIR~\citep{liang2021swinir}&ICCVW'21&21.165 &0.8450&0.3995\\

Restormer~\citep{Zamir2021Restormer}&CVPR'22&21.536 & 0.8437 &0.3608\\

Uformer~\citep{wang2021uformer}&CVPR'22&21.303  & 0.8233   &0.4013\\

LLformer~\citep{LLformer}&AAAI'23&24.065&  0.8580  &0.3516\\

UHDFour~\citep{Li2023ICLR_uhdfour}&ICLR'23&26.226 & 0.9000  &\underline{0.2194}\\
LMAR~\citep{yu2024empowering}&CVPR'24&26.821&  0.8840 & 0.3076\\
UHDformer~\citep{uhdformer} &AAAI'24&\underline{27.113}& \underline{0.9271} &0.2240\\
\textbf{UHDPromer (\textit{Ours})} &-&\textbf{27.159}&  \textbf{0.9285}    & \textbf{0.2118}
%0.7430M
\\\Xhline{1.5pt}
\end{tabular}
%\vspace{-2mm}
\end{table*}

\begin{figure*}[!t]
\centering
\begin{center}
\begin{tabular}{cccc}
\hspace{-2mm}\includegraphics[width=1\linewidth]{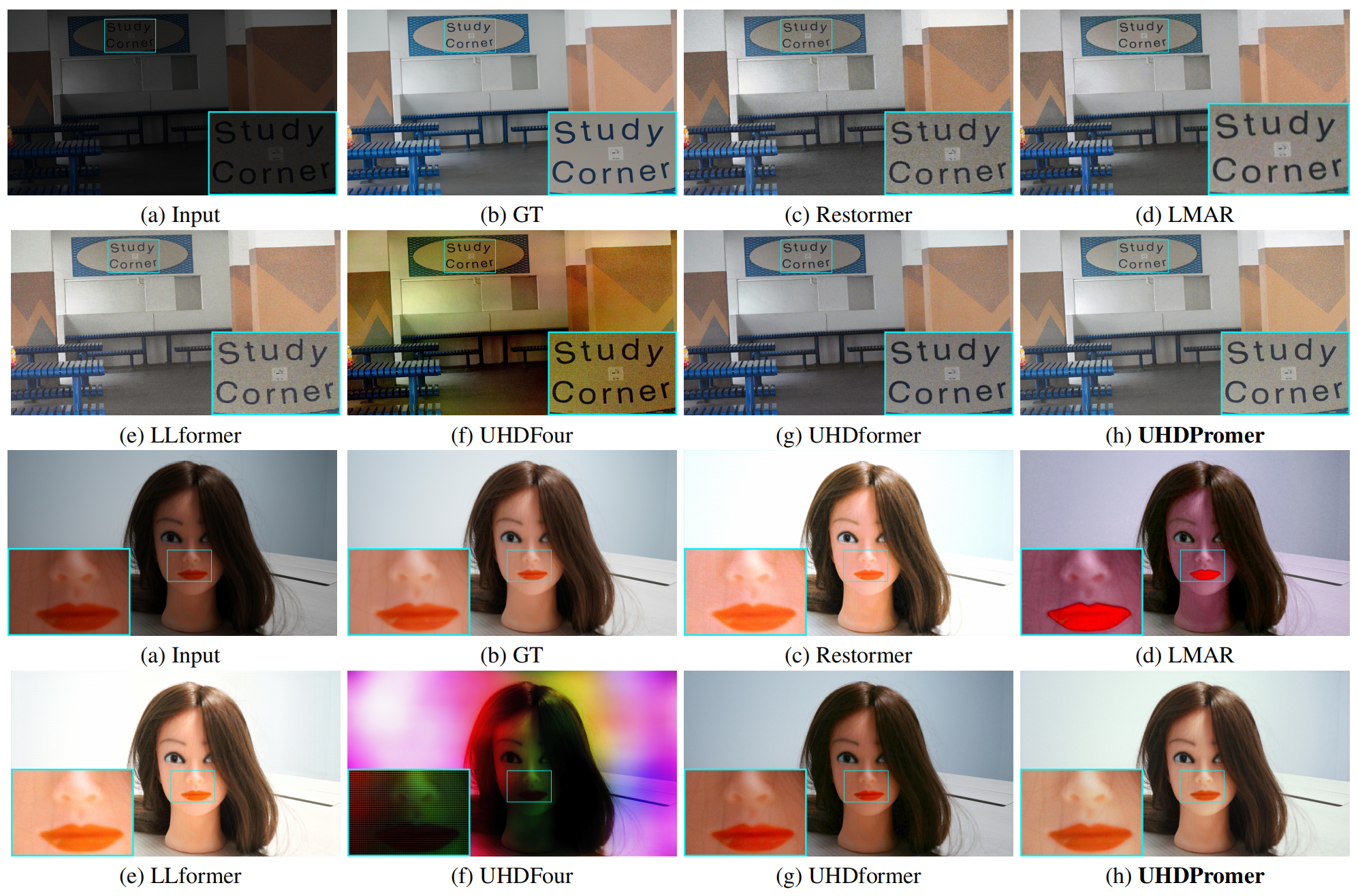} 
% \\
 
\end{tabular}
% \vspace{-1mm}
\caption{\textbf{UHD low-light enhancement on UHD-LL} \citep{Li2023ICLR_uhdfour} under \textcolor{blue}{\textit{\textbf{Setting 1}}}.
\textbf{UHDPromer} is able to produce clearer results with vivid colors.
%
% Best viewed with zoom-in.
}
\label{fig:Visual comparisons of low-light enhancement on UHD-LL under setting 1}
\end{center}
\end{figure*}
\subsubsection{Benchmark Datasets and Evaluation Settings}
Following the experimental protocol established by UHDformer~\citep{uhdformer}, we evaluate our method on three UHD datasets: UHD-LL~\citep{Li2023ICLR_uhdfour} for UHD low-light image enhancement, UHD-Haze for UHD image dehazing, and UHD-Blur for UHD image deblurring.
UHD-LL~\citep{Li2023ICLR_uhdfour} consists of paired low-light/normal-light images captured from real-world scenes with minimal noise artifacts. For UHD-Haze and UHD-Blur, we adopt the reorganized versions provided by~\citep{uhdformer}, which were reconstructed from the original datasets~\citep{Zheng_uhd_CVPR21,Deng_2021_ICCV_uhd_denlurring} due to file accessibility issues in the original releases. 
These reorganized datasets maintain the essential characteristics of the original collections while ensuring data integrity and accessibility.
The dataset statistics of UHD image restoration benchmarks are summarised as Tab.~\ref{tab: Datasets Statistics.}.
Following~\citep{uhdformer}, when evaluating the restoration performance, there are two settings:
\begin{description}
% \vspace{-2mm}
\item[\textcolor{blue}{\textit{\textbf{Setting 1:}}}] We exploit general image restoration datasets to train deep models and then test on UHD testing images in UHD-LL, UHD-Haze, and UHD-Blur.
In this study, we use widely-used LOL~\citep{retinexnet_wei_bmvc18}, SOTS-ITS~\citep{RESIDE_dehazingbenchmarking_tip2019}, and GoPro~\citep{gopro2017} as general low-light image enhancement, dehazing, and deblurring benchmark datasets.
\item[\textcolor{blue}{\textit{\textbf{Setting 2:}}}] We train deep models on UHD restoration benchmarks, i.e., UHD-LL, UHD-Haze, and UHD-Blur training sets, then test on corresponding UHD testing datasets.
% \vspace{-3mm}
\end{description}
%%%%%%%%%%%%%%%%%%
%
\begin{figure*}[t]
% \scriptsize
\centering
\begin{center}
\begin{tabular}{cccc}
\hspace{-2mm}\includegraphics[width=1\linewidth]{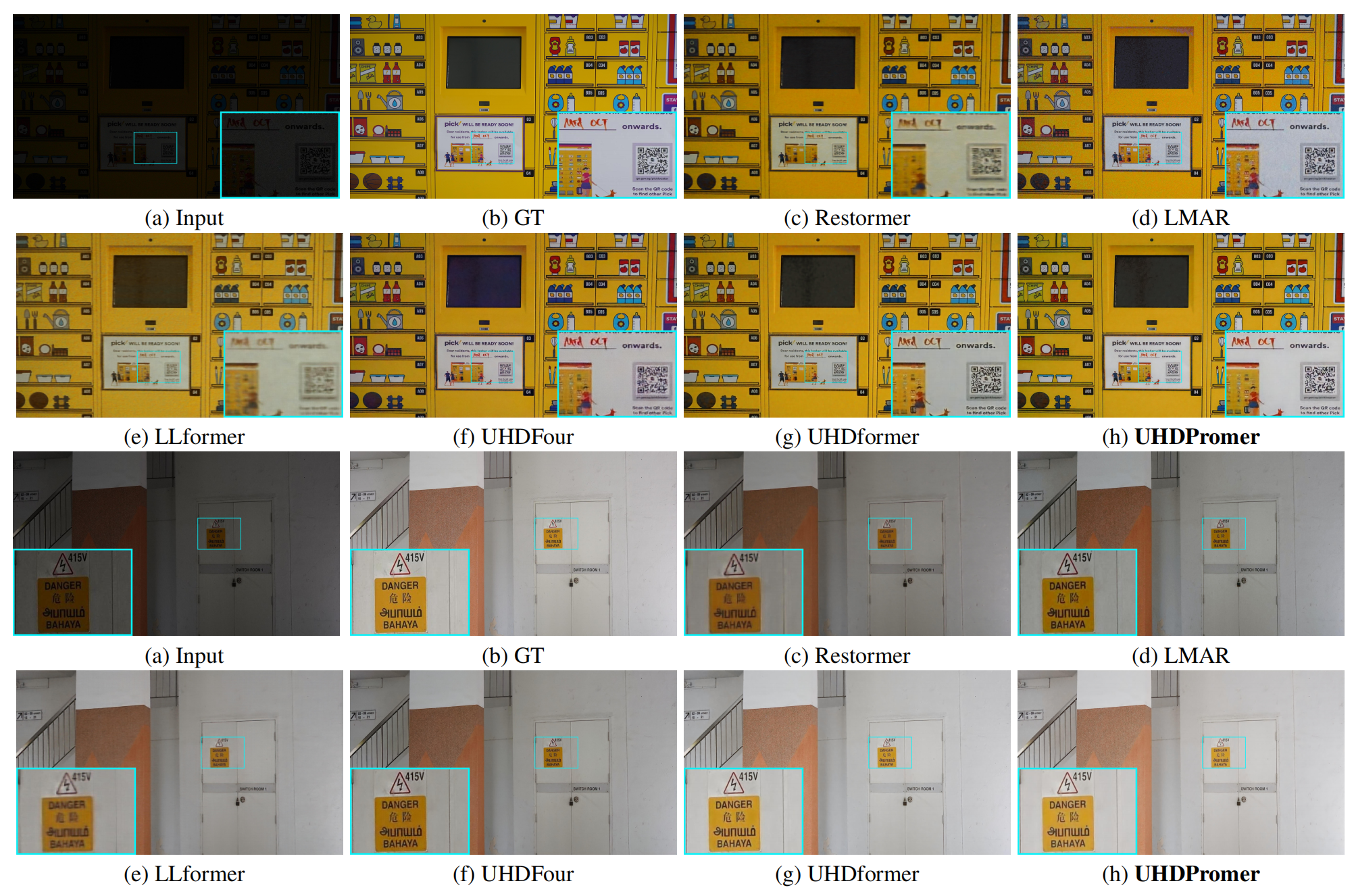} 
\end{tabular}
% \vspace{-1mm}
\caption{\textbf{UHD low-light enhancement on UHD-LL} \citep{Li2023ICLR_uhdfour} under \textcolor{blue}{\textit{\textbf{Setting 2}}}.
\textbf{UHDPromer} is capable of generating clearer results with more natural colors.
}
\label{fig:Visual comparisons of low-light enhancement on UHD-LL under setting 2.}
\end{center}
% \vspace{-3mm}
\end{figure*}
%
% \\
% \noindent \textbf{Evaluation Metrics.}
\subsubsection{Evaluation Metrics}
% \subsection{Evaluation Metrics}
%
Following UHDFour~\citep{Li2023ICLR_uhdfour}, we adopt commonly-used IQA PyTorch Toolbox~\footnote{ https://github.com/chaofengc/iqa-pytorch} to compute the PSNR~\citep{PSNR_thu} and SSIM~\citep{SSIM_wang} scores.
We also evaluate model sizes by reporting the learnable parameters (Param).
Since some methods cannot directly process full-resolution UHD images, we have to adopt an additional method to conduct the experiments.
According to UHDFour~\citep{Li2023ICLR_uhdfour}, resizing the input to the largest size manageable by the model yields better results than dividing the input into four patches and subsequently stitching the result.
\textbf{\subsection{Main Results}}
We qualitatively and quantitatively evaluate the UHD image restoration and enhancement results on UHD low-light image enhancement, UHD image dehazing, and UHD image deblurring.
%
% \noindent \textbf{Results on Low-Light Image Enhancement.} 
%
\begin{table*}[t]
% \scriptsize
%\vspace{-2mm}
\caption{\textbf{UHD image dehazing}. 
Our \textbf{UHDPromer} outperforms SOTA approaches.
}
\label{tab:Image dehazing.} 
%\tablestyle{25.25pt}{1.05}
\setlength{\tabcolsep}{25.25pt}
\begin{tabular}{l|c|cc|l}
\Xhline{1.5pt}
\textbf{Method} &\textbf{Venue} &   \textbf{PSNR}~$\uparrow$  & \textbf{SSIM}~$\uparrow$  & 
\textbf{LPIPS}~$\downarrow$  %&\textbf{Parameters}~$\downarrow$ ($\Delta_{\text{tiny}}$, $\Delta$)
\\
\Xhline{1pt}
 \multicolumn{5}{c}{\textcolor{blue}{\textit{\textbf{Setting 1:}}} Trained on SOTS-ITS}\\
\Xhline{1pt}
GridNet~\citep{grid_dehaze_liu}&ICCV'19&14.783& 0.8466&0.2518\\
MSBDN~\citep{msbdn_cvpr20_dong}&CVPR'20&15.043&\underline{ 0.8570}&0.2709\\
UHD~\citep{Zheng_uhd_CVPR21}&ICCV'21&11.708&0.6569&0.5948\\
Restormer~\citep{Zamir2021Restormer}&CVPR'22&13.875 & 0.6405 &0.4477\\
Uformer~\citep{wang2021uformer}&CVPR'22&15.264&  0.6724 & 0.4288\\
DehazeFormer~\citep{DehazeFormer}& TIP'23 &  14.721 &  0.6641  &0.4100\\
LMAR~\citep{yu2024empowering}&CVPR'24&\underline{16.896}&  0.7506 & 0.3741\\
UHDformer~\citep{uhdformer} &AAAI'24&15.325&0.8560  &\underline{0.2294}\\
\textbf{UHDPromer (\textit{Ours})} &-&\textbf{16.927}&  \textbf{0.8666}  &\textbf{0.2034}\\
%0.7430M\\
\Xhline{1pt}
 \multicolumn{5}{c}{\textcolor{blue}{\textit{\textbf{Setting 2:}}} Trained on UHD-Haze}\\
\Xhline{1pt}
UHD~\citep{Zheng_uhd_CVPR21}&ICCV'21&18.043 & 0.8113 &0.3593\\

Restormer~\citep{Zamir2021Restormer}&CVPR'22& 12.718 &  0.6930 &0.4560\\

Uformer~\citep{wang2021uformer}&CVPR'22& 19.828 &  0.7374 &0.4220\\

DehazeFormer~\citep{DehazeFormer}&TIP'23&14.986 &  0.7202  & 0.3998\\

LMAR~\citep{yu2024empowering}&CVPR'24&20.728&  0.9192& 0.1789\\
UHDformer~\citep{uhdformer} &AAAI'24&\underline{22.586}&\underline{0.9427}  &\underline{0.1188}\\
\textbf{UHDPromer (\textit{Ours})} &-&\textbf{22.725} & \textbf{0.9432}  & \textbf{0.1134}\\
%0.7430M\\
\Xhline{1.5pt}
\end{tabular}
% \vspace{-3mm}
% \vspace{-2mm}
\end{table*}
\begin{figure*}[!t]
% \scriptsize
\centering
\begin{center}
\begin{tabular}{cccc}
\hspace{-2mm}\includegraphics[width=01\linewidth]{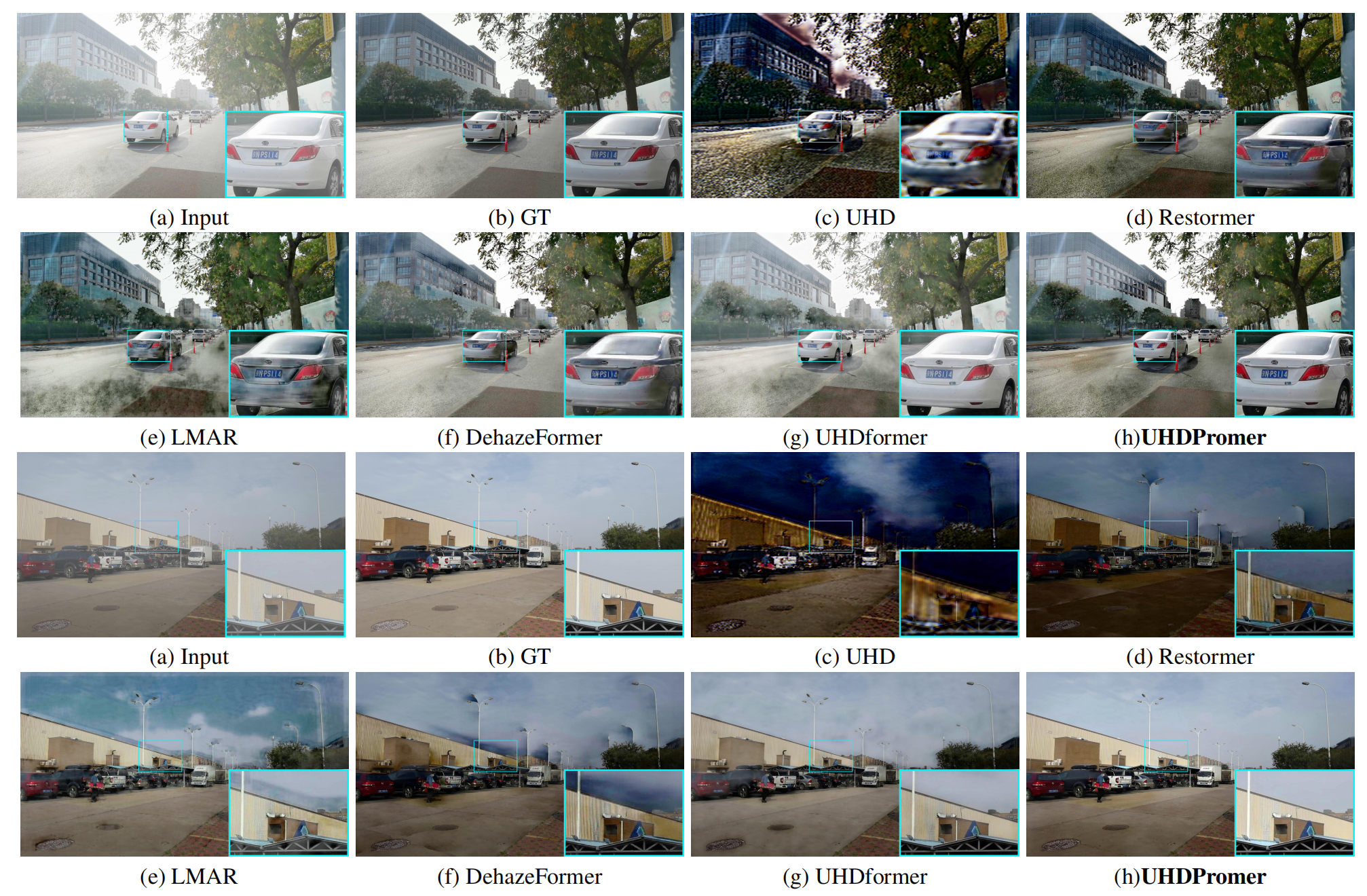} 
\end{tabular}
 \vspace{-2mm}
\caption{\textbf{UHD image dehazing on UHD-Haze} \citep{uhdformer} under \textcolor{blue}{\textit{\textbf{Setting 1}}}.
\textbf{UHDPromer} is able to generate much cleaner results.
%
% Best viewed with zoom-in.
}
\label{fig:Visual comparisons of image dehazing on UHD-Haze under setting 1.}
\end{center}
% \vspace{-7mm}
\end{figure*}
\begin{figure*}[!t]
% \scriptsize
\centering
\begin{center}
\begin{tabular}{ccccccccc}
\hspace{-2mm}\includegraphics[width=1\linewidth]{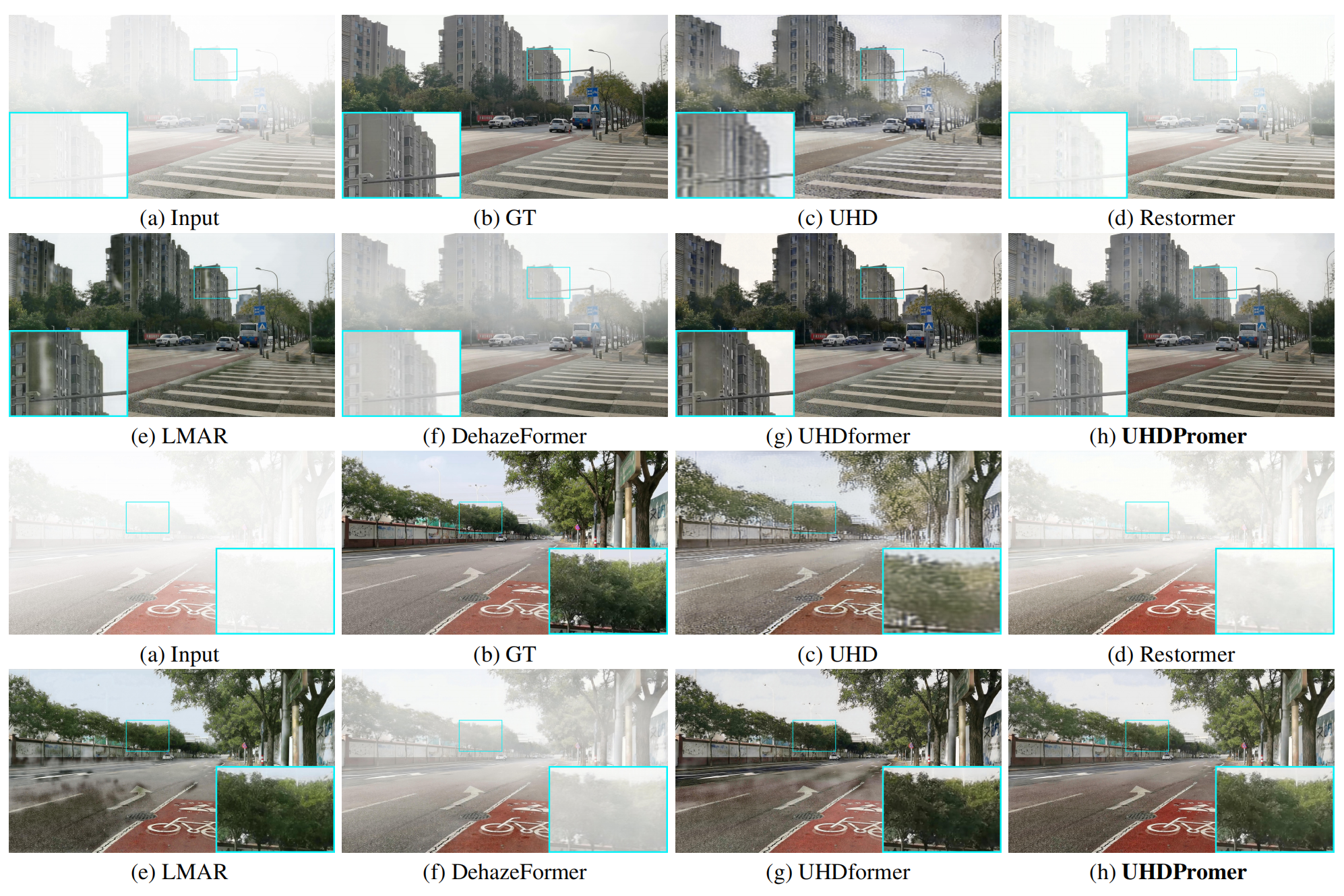} 
\end{tabular}
\vspace{-3mm}
\caption{\textbf{UHD dehazing on UHD-Haze} \citep{uhdformer} under \textcolor{blue}{\textit{\textbf{Setting 2}}}.
\textbf{UHDPromer} can generate much clearer dehazing results.
%
% Best viewed with zoom-in.
}
\label{fig:Visual comparisons of image dehazing on UHD-Haze under setting 2.}
\end{center}
\vspace{-4mm}
\end{figure*}
\subsubsection{Results on Low-Light Image Enhancement}
We evaluate UHD low-light image enhancement results under different training settings.
Tab.~\ref{tab:Low-light image enhancement.} shows that our UHDPromer advances state-of-the-art approaches in both of these two training settings. 
Compared with recent state-of-the-art UHDFour\footnote{Note that we use the publicly-released codes to re-train UHDFour~\citep{Li2023ICLR_uhdfour} with a shuffle down factor of 8 on LOL~\citep{retinexnet_wei_bmvc18} for fair comparisons with our method.}~\citep{Li2023ICLR_uhdfour}, our UHDPromer respectively achieves $6.943$dB and $0.933$dB PSNR gains
%saves at least 89$\times$ and 24$\times$ learnable parameters 
while consistently advancing it under different training settings.
Note that UHDFour under setting 1 cannot handle UHD images, while our UHDformer still keeps excellent enhancement performance.
Moreover, although some existing methods, e.g., UHDformer~\citep{uhdformer}
%\citep{zhao_lie} and \citep{Wu_2022_CVPR}
can handle full-resolution UHD images without other processing steps, our UHDPromer still outperforms them.
We further notice that the existing method, especially designed for image enhancement, LMAR~\cite{yu2024empowering}, significantly outperforms all approaches in terms of SSIM under setting 1 but exhibits inferior quality under setting 2.
This may indicate that LMAR is not robust enough under different training data.
Fig.~\ref{fig:Visual comparisons of low-light enhancement on UHD-LL under setting 1} and Fig.~\ref{fig:Visual comparisons of low-light enhancement on UHD-LL under setting 2.} respectively present the visual comparisons on UHD-LL under different settings, where UHDPromer is able to generate clearer results with more natural colors.
% \\
% \noindent \textbf{Results on Image Dehazing.}
\subsubsection{Results on Image Dehazing}
Tab.~\ref{tab:Image dehazing.} summarises the quantitative dehazing results with different training settings.  
Compared to recent work DehazeFormer~\citep{DehazeFormer} and UHDformer~\citep{uhdformer}, our UHDPromer respectively 
%reduces 129$\times$ and 34$\times$ model sizes while 
achieves $7.739$dB and $0.139$dB PSNR gains under setting 2. 
We notice that although some CNN-based approaches, e.g., GridNet~\citep{grid_dehaze_liu}, UHD~\citep{Zheng_uhd_CVPR21}, and MSBDN~\citep{msbdn_cvpr20_dong}, can directly handle UHD images, they are less effective in handling the UHD images under setting 1, while our UHDformer still keeps state-of-the-art performance.
Further, we note that Restormer's~\citep{Zamir2021Restormer} metrics under setting 2 are lower than cross-validation under setting 1.
This may indicate that Restormer is not effective and suitable to handle UHD hazes.
In contrast, our UHDPromer is more suitable to handle UHD images.
Fig.~\ref{fig:Visual comparisons of image dehazing on UHD-Haze under setting 1.} and Fig.~\ref{fig:Visual comparisons of image dehazing on UHD-Haze under setting 2.} present visual comparisons under different training scenarios, demonstrating that our UHDPromer is capable of producing clearer results. In contrast, alternative methods consistently result in either excessive haze or the creation of significant artifacts.
\begin{table*}[t]
% \scriptsize
% \vspace{-2mm}
\caption{\textbf{Image deblurring}. 
\textbf{UHDformer} outperforms SOTAs.
}
%\vspace{-4mm}
\label{tab:Image deblurring.} 
% \tablestyle{25.25pt}{1.05}
\setlength{\tabcolsep}{25.25pt}
\begin{tabular}{l|c|cc|l}
\Xhline{1.5pt}
\textbf{Method} &\textbf{Venue} &   \textbf{PSNR}~$\uparrow$  & \textbf{SSIM}~$\uparrow$  & 
\textbf{LPIPS}~$\downarrow$   %\textbf{Parameters}~$\downarrow$ ($\Delta_{\text{tiny}}$, $\Delta$)
\\
\Xhline{1pt}
 \multicolumn{5}{c}{\textcolor{blue}{\textit{\textbf{Setting 1:}}} Trained on GoPro}\\
\Xhline{1pt}
DMPHN~\citep{dmphn2019}&CVPR'19&26.490 &0.7985&\underline{0.2696}\\

MIMO-Unet++~\citep{cho2021rethinking_mimo}&ICCV'21&24.290 &0.7354&0.3723\\ 

MPRNet~\citep{Zamir_2021_CVPR_mprnet}&CVPR'21&24.571&0.7426&0.3597\\

Restormer~\citep{Zamir2021Restormer}&CVPR'22&24.872 & 0.7484& 0.3550\\

Uformer~\citep{wang2021uformer}&CVPR'22&24.382 & 0.7209 &0.3901\\

Stripformer~\citep{Tsai2022Stripformer}&ECCV'22&24.915 &0.7463&0.3299\\

FFTformer~\citep{Kong_2023_CVPR_fftformer}&CVPR'23&24.625 & 0.7396& 0.3428\\
LMAR~\citep{yu2024empowering}&CVPR'24&23.040&  0.7894 & 0.3513\\
UHDformer~\citep{uhdformer} &AAAI'24&\underline{27.436}& \underline{0.8231}& 0.2774\\
\textbf{UHDPromer (\textit{Ours})} &-&\textbf{27.582}&  \textbf{0.8263}  & \textbf{0.2554}\\
%0.7430M\\
\Xhline{1pt}
 \multicolumn{5}{c}{\textcolor{blue}{\textit{\textbf{Setting 2:}}} Trained on UHD-Blur}\\
\Xhline{1pt}
MIMO-Unet++~\citep{cho2021rethinking_mimo}&ICCV'21&25.025 &0.7517&0.3874\\

Restormer~\citep{Zamir2021Restormer}&CVPR'22&25.210&  0.7522&0.3695\\

Uformer~\citep{wang2021uformer}&CVPR'22&25.267&  0.7515 &0.3851\\

Stripformer~\citep{Tsai2022Stripformer}&ECCV'22&25.052&0.7501&0.3740\\

FFTformer~\citep{Kong_2023_CVPR_fftformer}&CVPR'23&25.409 & 0.7571 &0.3708\\
LMAR~\citep{yu2024empowering}&CVPR'24&26.472&  0.7989 &0.2470\\
UHDformer~\citep{uhdformer} &AAAI'24&\underline{28.821}& \underline{0.8440}&\underline{0.2350}\\
\textbf{UHDPromer (\textit{Ours})} &-&\textbf{29.527}&  \textbf{0.8584} &\textbf{0.2163}\\
%0.7430M\\
\Xhline{1.5pt}
\end{tabular}
% \vspace{-3mm}
% \vspace{-2mm}
\end{table*}
\begin{figure*}[!t]
% \scriptsize
% \vspace{-5mm}
\centering
\begin{center}
\begin{tabular}{cccc}
\hspace{-2mm}\includegraphics[width=1\linewidth]{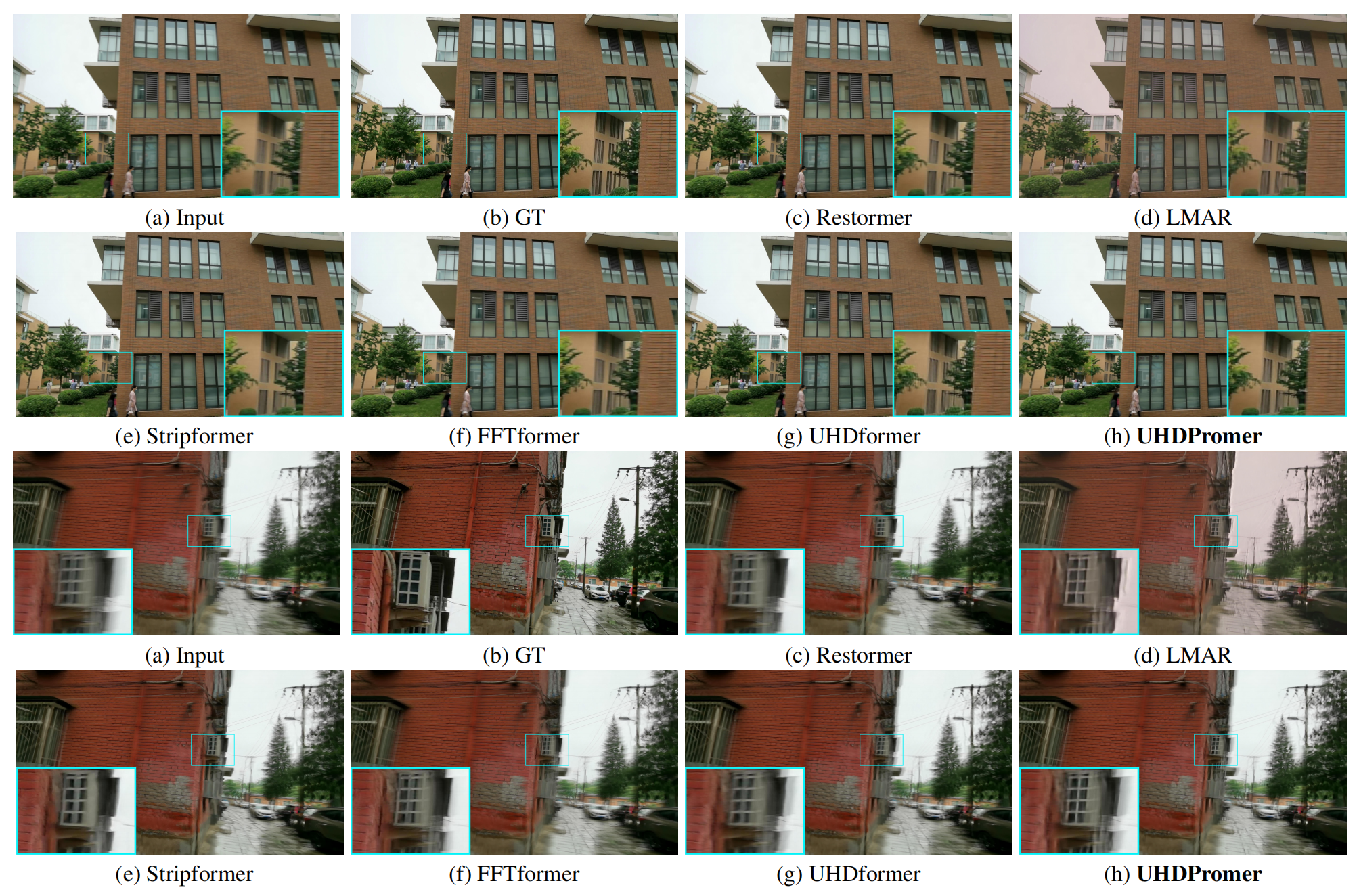} 
\end{tabular} % VID_010_00355_
\vspace{-3mm}
\caption{\textbf{UHD image deblurring on UHD-Blur} \citep{uhdformer} under \textcolor{blue}{\textit{\textbf{Setting 1}}}.
\textbf{UHDformer} is able to generate results with finer structures.
%
% Best viewed with zoom-in.
%
% Best viewed with zoom-in.
}
\label{fig:Image deblurring on UHD-Blur under setting 1.}
\end{center}
\vspace{-3mm}
\end{figure*}
\begin{figure*}[!t]
% \scriptsize
% \vspace{-6mm}
\centering
\begin{center}
\begin{tabular}{cccc}
\hspace{-2mm}\includegraphics[width=1\linewidth]{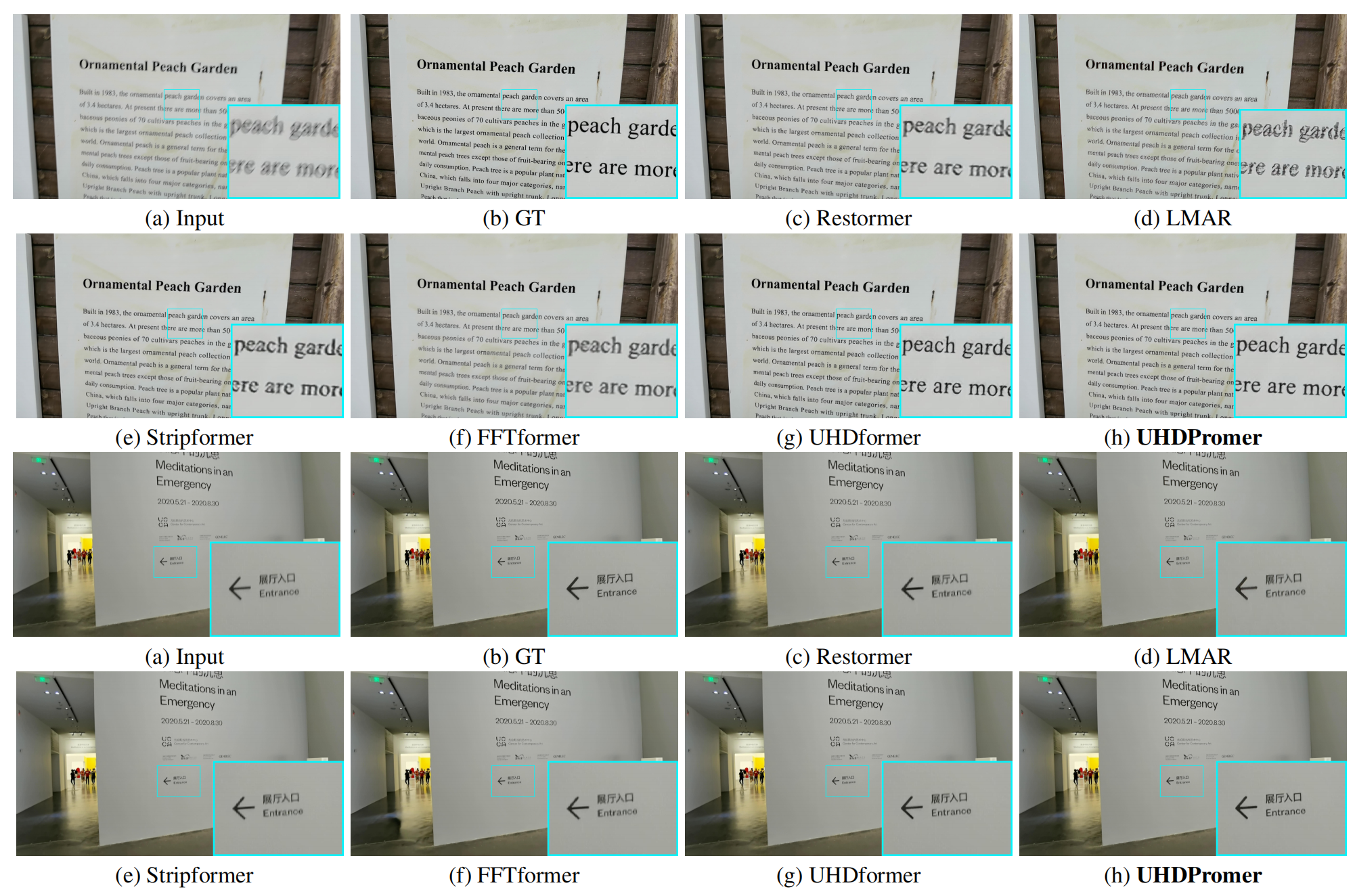} 
\end{tabular} % VID_010_00355_

\caption{\textbf{UHD image deblurring on UHD-Blur} \citep{uhdformer} under \textcolor{blue}{\textit{\textbf{Setting 2}}}.
\textbf{UHDformer} is capable of producing results with much clearer characters.
}
\label{fig:Image deblurring on UHD-Blur under setting 2.}
\end{center}
% \vspace{-4mm}
\end{figure*}
% \\
% \noindent \textbf{Results on Image Deblurring.} 
\subsubsection{Results on Image Deblurring}
We evaluate UHD image deblurring performance with two suggested training settings.
Tab.~\ref{tab:Image deblurring.} shows that our UHDPromer significantly advances state-of-the-art approaches under different training settings. 
Compared with FFTformer~\citep{Kong_2023_CVPR_fftformer} and UHDformer~\citep{uhdformer} which is the newest state-of-the-art deblurring approach, our UHDPromer respectively obtains $4.118$dB and $0.706$dB PSNR gains under setting 2.
We also note that some CNN-based methods, i.e., DMPHN~\citep{dmphn2019}, MIMO-Unet++~\citep{cho2021rethinking_mimo}, and MPRNet~\citep{Zamir_2021_CVPR_mprnet}, can handle the full-resolution UHD images, they consume about $100 \times$ and $30 \times$ more training parameters while dropping at least $0.774$dB and $1.092$dB PSNR under setting 1.
Fig.~\ref{fig:Image deblurring on UHD-Blur under setting 1.} and Fig.~\ref{fig:Image deblurring on UHD-Blur under setting 2.} respectively provide visual UHD deblurring examples, where our UHDPromer is able to produce sharper deblurring results.
% while existing state-of-the-art approaches always cannot handle the UHD images well.
%
\textbf{\subsection{Ablation Study}}
We use the UHD-LL~\citep{Li2023ICLR_uhdfour} to conduct the ablation study to analyze the proposed component and demonstrate the effectiveness of the proposed methods.
Except for reporting the final restoration performance (\textbf{Main Branch}), we also provide PSNR/SSIM scores in the super-resolution branch (\textbf{SR Branch}) for reference.
We also report the number of parameters of different models for reference in the ablation study.
\subsubsection{Effect on Neural Discrimination-Prompted Transformers}
A pivotal aspect of our UHDPromer is the introduction of the Neural Discrimination-Prompted Transformer (NDPT), guided by Neural Discrimination Priors (NDP). 
Therefore, an analysis exploring the effect of NDPT is imperative.
As revealed in Tab.~\ref{tab:Ablation study on NDPTB}, disabling NDP in either NDPA or NDPN, or in both, results in a noticeable performance degradation (Tab.~\ref{tab:Ablation study on NDPTB}(f) vs. Tab.~\ref{tab:Ablation study on NDPTB}(a), (b), and (c)).
For those curious about the implications of utilizing direct features, which means the outputs of stride convolutions, we substitute the NDP with these direct features. 
We note that the performance declines when incorporating NDP into NDPN.
This may be because integrating NDP into NDPA makes Transformers pay attention to discriminative content.
However, solely using NDPN may not pay attention to more discriminative features, so that the NDPN cannot effectively handle intermediate features, leading to a declined in performance.

\begin{figure}[!t]
% \scriptsize
% \vspace{-8mm}
    % \vspace{0pt}
\centering
\begin{tabular}{ccccccccc}
\hspace{-2mm}\includegraphics[width=1\linewidth]{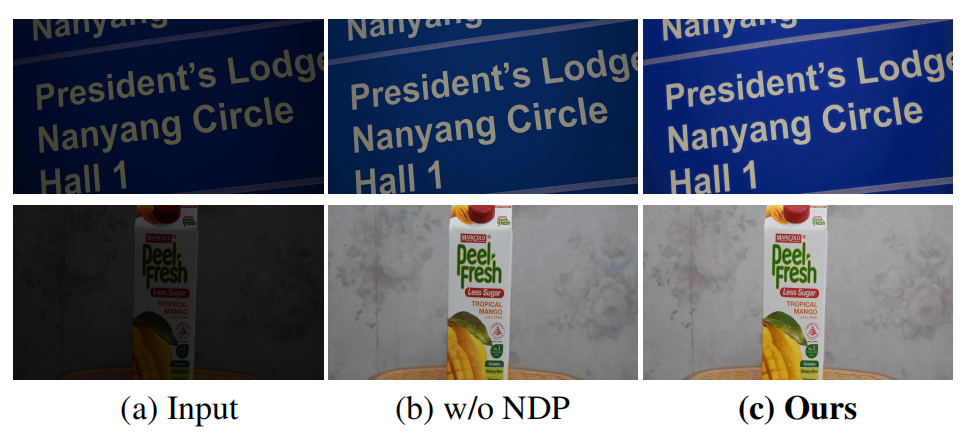} 

\end{tabular} 
% \vspace{-3mm}
\caption{\textbf{Visual effect on neural discrimination priors.}
Using NDP is capable of helping produce visually pleasing results with more natural colors.
}
\label{fig:Visual effect on NDPL.}
%\vspace{-3mm}
\end{figure}

Tab.~\ref{tab:Ablation study on NDPTB} illustrates that the employment of our refined neural discrimination priors leads to a superior model, achieving a gain of $1.021$dB PSNR over the model utilizing direct features (Tab.~\ref{tab:Ablation study on NDPTB}(f) vs. Tab.~\ref{tab:Ablation study on NDPTB}(d)).
To further elucidate the efficacy of our NDP within the attention and forward-network mechanisms, we introduce NDP before each Transformer block, as opposed to their integration solely in the attention and forward-network designs. This approach manifested in approximately $1$dB PSNR improvement (Tab.~\ref{tab:Ablation study on NDPTB}(f) vs. Tab.~\ref{tab:Ablation study on NDPTB}(e)).
Fig.~\ref{fig:Visual effect on NDPL.} provides visual examples, where using NDP to guide the learning of Transformers helps produce visually pleasing results with more natural colors.
\begin{table*}[t]
% \scriptsize
%\vspace{-2mm}
\caption{\textbf{Ablation study on neural discrimination-prompted Transformers}.
Inserting NDP in both attention and forward-network produces a positive effect.
}
% \vspace{-2mm}
\label{tab:Ablation study on NDPTB} 
% \tablestyle{16pt}{1.05}
\setlength{\tabcolsep}{16pt}
\begin{tabular}{l|l|cc|cc|cc}
\Xhline{1.5pt}
\multirow{2}{*}{\textbf{ID}} & \multirow{2}{*}{\textbf{Experiment}} & \multicolumn{2}{c|}{\textbf{SR Branch}} & \multicolumn{2}{c|}{\textbf{Main Branch}}  & \multirow{2}{*}{\textbf{Parameters}~$\downarrow$}
\\
&& \textbf{PSNR}~$\uparrow$ &\textbf{SSIM}~$\uparrow$ & \textbf{PSNR}~$\uparrow$ & \textbf{SSIM}~$\uparrow$
\\
\Xhline{1pt}
\textcolor{blue}{\textbf{(a)}}& w/o NDP in NDPA\&NDPN& 26.487 & 0.8974&26.811&0.9282&0.5322M \\
\textcolor{blue}{\textbf{(b)}}& w/o NDP in NDPA&25.962 &0.8952 & 26.183&0.9252&0.7348M\\
\textcolor{blue}{\textbf{(c)}}& w/o NDP in NDPN&26.745 &0.8984 &27.026&0.9283 &0.5404M\\
\textcolor{blue}{\textbf{(d)}}& NDP$\rightarrow$ Direct Feature &25.999 &\textbf{0.9021}& 26.138&0.9272&0.7430M  \\
\textcolor{blue}{\textbf{(e)}}& Using NDP before NDPTB &25.790  &0.8974& 26.161& 0.9263&0.5322M \\
% \cdashline{1-6}[3pt/2.5pt]
\textcolor{blue}{\textbf{(f)}}& \textbf{Full Model} (\textit{\textbf{Ours}})&\textbf{26.859} &0.8985 &\textbf{27.159} &\textbf{0.9285} &0.7430M \\
\Xhline{1.5pt}
\end{tabular}
% \vspace{-3mm}
% \vspace{-2mm}
\end{table*}
\begin{table*}[!t]
% \scriptsize
%\vspace{-2mm}
\caption{\textbf{Effect on super-resolution-guided reconstruction}.
The introduced SR-guided reconstruction significantly improves restoration quality in terms of SSIM.
}
\label{tab:Effect on super-resolution-guided reconstruction} 
% \vspace{-2mm}
\centering
% \tablestyle{16.75pt}{1.05}
\setlength{\tabcolsep}{16.75pt}
\begin{tabular}{l|l|cc|cc|cc}
\Xhline{1.5pt}
\multirow{2}{*}{\textbf{ID}} & \multirow{2}{*}{\textbf{Experiment}} & \multicolumn{2}{c|}{\textbf{SR Branch}} & \multicolumn{2}{c|}{\textbf{Main Branch}}  & \multirow{2}{*}{\textbf{Parameters}~$\downarrow$}
\\
&& \textbf{PSNR}~$\uparrow$ & \textbf{SSIM}~$\uparrow$ & \textbf{PSNR}~$\uparrow$ & \textbf{SSIM}~$\uparrow$
\\
\Xhline{1pt}
\textcolor{blue}{\textbf{(a)}}& Cascaded Reconstruction& - &-  &26.375&0.9043&0.7414M  \\
\textcolor{blue}{\textbf{(b)}}& w/o Second Term in \eqref{eq: loss}&- &-&27.125&0.9051 &0.7425M \\
% \cdashline{1-6}[3pt/2.5pt]
\textcolor{blue}{\textbf{(c)}}& \textbf{w/ SR Branch (\textit{Ours})}&\textbf{26.859} &\textbf{0.8985} & \textbf{27.159}&\textbf{0.9285}&0.7430M\\
\Xhline{1.5pt}
\end{tabular}
% \vspace{-3mm}
% \vspace{-2mm}
\end{table*}
\begin{figure}[!t]
% \scriptsize
\centering
\begin{tabular}{ccccccccc}
\hspace{-2mm}\includegraphics[width=1\linewidth]{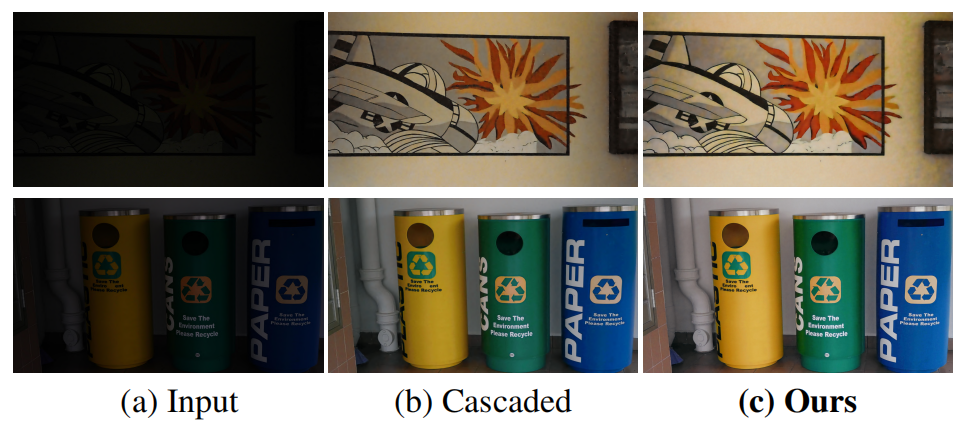} 
\end{tabular} 
% \vspace{-2mm}
\caption{\textbf{Visual effect on SR-guided reconstruction.}
The proposed SR-guided reconstruction is able to improve the visual results with more vivid colors.
}
\label{fig:Visual effect on SR-guided reconstruction.}
% \vspace{-2mm}
\end{figure}
\begin{figure}[!t]
% \vspace{-5mm}
% \scriptsize
%\vspace{-10mm}
\centering
\includegraphics[width = 1\linewidth]{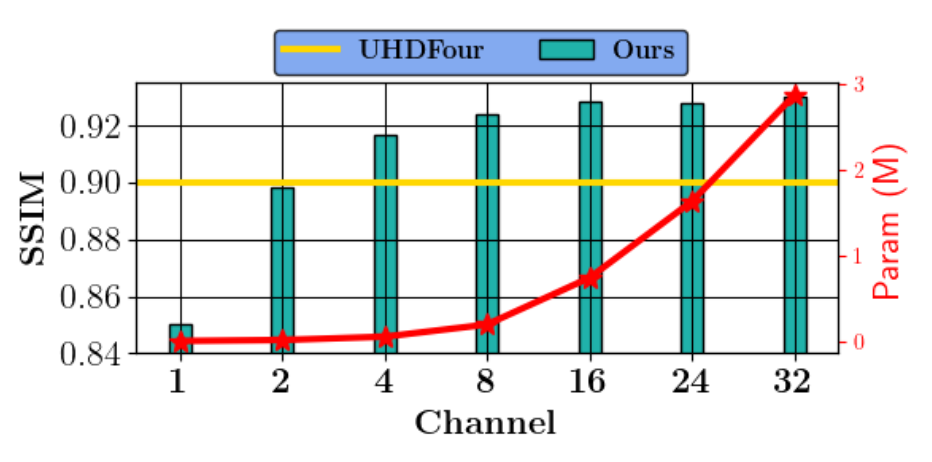}
\vspace{-5mm}
\caption{\textbf{Effect on number of channels}.
Our model, with only $2$ channels and 0.0168M parameters, achieves a performance comparable to UHDFour~\citep{Li2023ICLR_uhdfour}, while being approximately \textit{\textbf{1,000 times smaller}} in model size (UHDFour has $17.5$M parameters, as seen in Tab.~\ref{tab: Comparisons on Parameters.}).
}
\label{fig: Effect on number of channels.}
%\vspace{-2mm}
%\end{wrapfigure}
\end{figure}
\begin{figure}[!t]
% \vspace{-4mm}
\begin{center}
\begin{tabular}{cccccccccc}
\includegraphics[width = 1\linewidth]{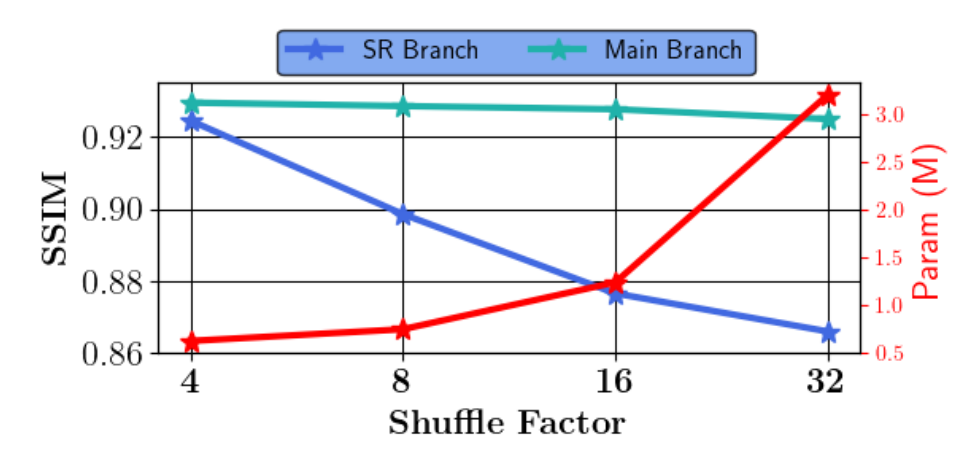}
\end{tabular}
\vspace{-5mm}
\caption{\textbf{Effect on shuffle down factor}.
It is noteworthy that while the model using a shuffle down factor of 4 achieves a larger receptive field compared to the baseline model with a factor of 8, there is no significant improvement in SSIM.
}
\label{fig: Effect on Shuffled Down Factor.}
\end{center}
% \vspace{-5mm}
\end{figure}
\subsubsection{Effect on Super-Resolution-Guided Reconstruction}
A pivotal innovation in our UHDPromer is the integration of a super-resolution-guided reconstruction. To assess its impact, we have carried out two additional experiments.
In the first experiment, we omit the super-resolution branch, channeling the shuffle up features directly to the main branch, and utilize $5$ ConvNeXt-v2 blocks~\citep{woo2023convnext} (whose number is equal to the total number of ConvNeXt-v2 in both the main branch and the SR branch) for the reconstruction of final results. 

\begin{table*}[!t]
\caption{\textbf{Parameters comparisons}. %on $1024$$\times$$1024$ pixels.
Our UHDPromer significantly reduces the number of parameters compared to the advanced methods, but is slightly above UHDformer.
}
\label{tab: Comparisons on Parameters.} 
% \vspace{-3mm}
\centering
% \tablestyle{0.9pt}{1.05}
\setlength{\tabcolsep}{0.9pt}
\begin{tabular}{l|ccccccc}
\Xhline{1.5pt}
\multirow{2}{*}{\textbf{Method}}  & LMAR& UHD& UHDFour& FFTformer& DehazeFormer& UHDformer& \textbf{UHDPromer} 
\\
& \citep{yu2024empowering}& \citep{Zheng_uhd_CVPR21}& \citep{Li2023ICLR_uhdfour}& \citep{Kong_2023_CVPR_fftformer}& \citep{DehazeFormer}
&\citep{uhdformer}
& (\textbf{\textit{Ours}})
\\
\Xhline{1pt}
 \textbf{Parameters (M)} &1.314&34.5&17.5&16.6&2.5& \textbf{0.3393} &\underline{0.7430}
\\
% \cdashline{1-7}[3pt/2.5pt]
 \textbf{Reduction} &43.5\%&97.8\%&95.8\%&95.5\%&70.3\%&119.0\% &-
\\
\Xhline{1.5pt}
\end{tabular}
\end{table*}
\begin{table*}[!t]
\caption{\textbf{FLOPs comparisons} on $1024$$\times$$1024$ pixels.
Our UHDPromer significantly reduces the computational complexity in terms of FLOPs, indicating the efficiency of our approach.
}
\label{tab: Comparisons on FLOPs.} 
% \vspace{-2mm}
\centering
% \tablestyle{2pt}{1.05}
\setlength{\tabcolsep}{2pt}
\begin{tabular}{l|ccccccc}
\Xhline{1.5pt}
\multirow{2}{*}{\textbf{Method}}  & LMAR& UHD& UHDFour& FFTformer& DehazeFormer& UHDformer& \textbf{UHDPromer} 
\\
& \citep{yu2024empowering}& \citep{Zheng_uhd_CVPR21}& \citep{Li2023ICLR_uhdfour}& \citep{Kong_2023_CVPR_fftformer}& \citep{DehazeFormer}
&\citep{uhdformer}& 
(\textbf{\textit{Ours}})
\\
\Xhline{1pt}
 \textbf{FLOPs (G)} &208.34&113.46&75.63&2107.94&375.40&\underline{51.63} &\textbf{32.56}
\\
% \cdashline{1-7}[3pt/2.5pt]
 \textbf{Reduction} &84.4\%&71.3\%&56.9\%&98.5\%&91.3\%&36.9\% &-
\\
\Xhline{1.5pt}
\end{tabular}
%\vspace{-2mm}
\end{table*}
\begin{table*}[!t]% \scriptsize
%\vspace{-6mm}
\caption{\textbf{Running Time (RT)} ($1024 \times 1024$ pixels) on the same GPU. 
Our \textbf{UHDPromer} runs faster than Transformer-based methods \textcolor{blue}{as well as the CNN-based approach~\citep{yu2024empowering}.} %UHDFour but faster than the UHD approach.
}
\label{tab: Running time comparisons.} 
% \vspace{-2mm}
\centering
% \tablestyle{7.3pt}{1.05}
\setlength{\tabcolsep}{7.3pt}
\begin{tabular}{l|ccccccccc}
\Xhline{1.5pt}
% \shline
\multirow{2}{*}{\textbf{Method}} &Restormer&LMAR&DehazeFormer&FFTformer&UHDformer&\textbf{UHDPromer}
\\
 &\citep{Zamir2021Restormer}&\citep{yu2024empowering}&\citep{DehazeFormer}& \citep{Kong_2023_CVPR_fftformer}
&\citep{uhdformer}&(\textbf{\textit{Ours}})
\\
\Xhline{1pt}
\textbf{ RT (s)}
 &1.86&3.40 &0.45 &1.27 &0.16 &\textbf{0.12}
%&0.173&\textbf{0.014}&0.445&0.741&0.807&&\underline{0.040}
\\
\Xhline{1.5pt}
\end{tabular}
% \vspace{-4mm}
\end{table*}

This approach, known as cascaded reconstruction, guarantees equitable comparisons.
In the second experiment, the super-resolution loss, denoted as the second term in \eqref{eq: loss}, is not utilized.
Tab.~\ref{tab:Effect on super-resolution-guided reconstruction} reveals that our super-resolution-guided reconstruction attains a $0.784$dB PSNR improvement compared to the model employing cascaded reconstruction while maintaining a similar parameter count. 
Notably, the omission of the SR loss, i.e., the second term in \eqref{eq: loss}, results in a decline in performance.
Fig.~\ref{fig:Visual effect on SR-guided reconstruction.} presents visual results, where our proposed SR-guided reconstruction yields images with heightened color vibrancy.

\subsubsection{Effect on Number of Channels}
We also examine the influence of the number of channels on performance, as depicted in Fig.~\ref{fig: Effect on number of channels.}. 
Additionally, we reference the performance of UHDFour~\citep{Li2023ICLR_uhdfour} to establish the robust baseline of our UHDPromer.
Fig.~\ref{fig: Effect on number of channels.} illustrates that there is a gradual enhancement in performance, in terms of SSIM, as the number of channels increases up to 16. Beyond 16 channels, no substantial gains in performance are observed.
It is noteworthy that our model, even with just $2$ channels, exhibits performance on par with UHDFour~\citep{Li2023ICLR_uhdfour} while consuming about \textit{\textbf{1,000}} smaller sizes compared to the model size of UHDFour, demonstrating the strong baseline of our UHDPromer for UHD image enhancement.
\subsubsection{Effect on Shuffle Down Factor}\label{sec: Effect on Shuffled Down Factor}

In our experiment, we employ a shuffle down factor of 8 to decrease the resolution, facilitating learning in a low-resolution space.
Additionally, we analyze the impacts of various other shuffle down factors, as illustrated in Fig.~\ref{fig: Effect on Shuffled Down Factor.}.
It is noteworthy that while the model using a shuffle down factor of 4 achieves a larger receptive field compared to the baseline model with a factor of 8, there is no significant improvement in SSIM.
In comparison, our model with a shuffle down factor of 8 demonstrates superior SSIM performance against models employing a larger shuffle down factor.

\begin{figure*}[!t]
%\vspace{2mm}
\centering
\begin{center}
\begin{tabular}{cccccc}
\hspace{-2mm} \includegraphics[width = 1\linewidth]{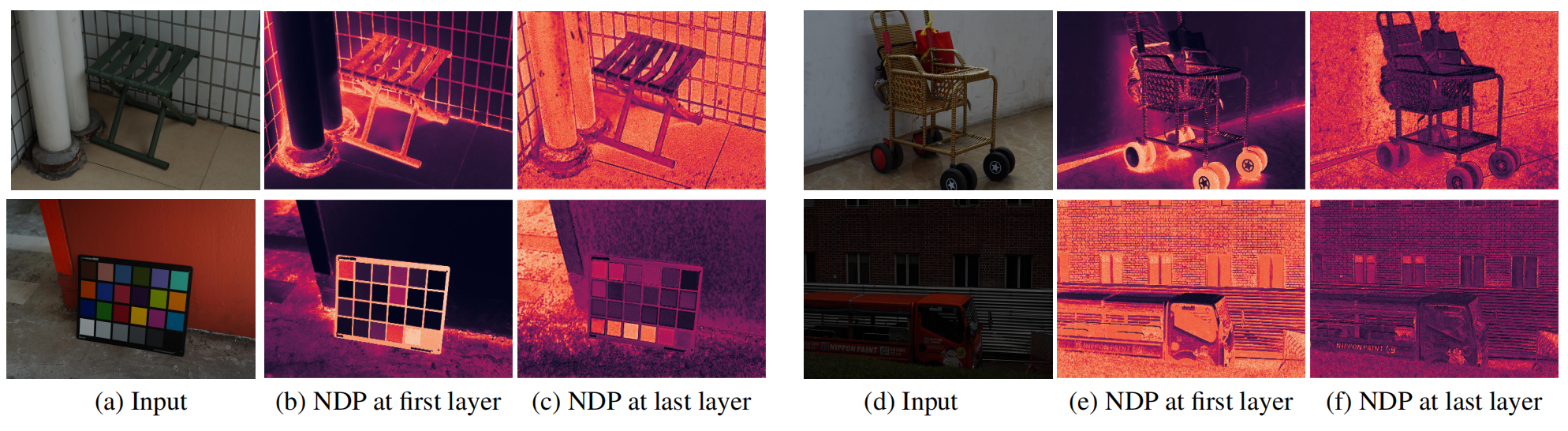} 
\end{tabular}
% \vspace{-3mm}
\caption{\textbf{Illustration of Neural Discrimination Priors} (NDP) at the first and last layer.
Our NDP can better measure the salient structure at both the first and last layer, which can serve as the discrimination function to help the learning of Transformers to facilitate restoration.
}
\label{fig:Illustration of Implicit Neural Discrimination Priors at the first layer}
\end{center}
% \vspace{-4mm}
\end{figure*}

%\vspace{-2mm}
%
\subsubsection{Demonstration on Neural Discrimination Priors}
From Fig.~\ref{fig:Illustration of Implicit Neural Discrimination Priors at the first layer}, we illustrate the Neural Discrimination Priors (NDP) at the first and last layers.
One can obviously observe that our NDP can effectively measure whether the structures lie in shallower layers or deeper layers, which demonstrates that the NDP is an effective prior, serving as the discrimination function to help learn more representative features in the low-resolution domain.

%

%\vspace{-4mm}
\subsubsection{Computational Complexity}
Tab.~\ref{tab: Comparisons on Parameters.} shows the parameter comparisons on several advanced methods. Our UHDPromer significantly reduces the number of parameters compared to some advanced methods and reduces at least 70.3\%, but is slightly above UHDformer.
Tab.~\ref{tab: Comparisons on FLOPs.} shows FLOPs comparisons.
Compared with Transformer-based methods, e.g., Restormer~\citep{Zamir2021Restormer}, FFTformer~\citep{Kong_2023_CVPR_fftformer}, and DehazeFormer~\citep{DehazeFormer}, our UHDPromer reduces at least 91.3\% FLOPs.
On the other hand, compared with UHD~\citep{Zheng_uhd_CVPR21}, UHDFour~\citep{Li2023ICLR_uhdfour}, LMAR~\cite{yu2024empowering}, and UHDformer~\citep{uhdformer}, where both of them are specifically designed for UHD images, our UHDPromer reduces at least 36.9\% FLOPs.
Tab.~\ref{tab: Running time comparisons.} shows running time (RT) comparisons, where our UHDPromer significantly runs faster than Transformer-based methods.
Notably, compared with the CNN-based approach LMAR~\citep{yu2024empowering}, our UHDPromer significantly outperforms it in terms of model size, computational complexity, and running time, further indicating the effectiveness and efficiency of our method.
%UHDFour~\citep{Li2023ICLR_uhdfour} but faster than the UHD approach~\citep{Zheng_uhd_CVPR21}.
%
% The results adequately indicates the efficiency of our approach.
%\vspace{-4mm}
\textbf{\subsection{Results on General Image Restoration and Enhancement}}\label{sec: Results on General Image Restoration}

Our UHDPromer is specifically designed for Ultra-High Definition (UHD) images, with the majority of its operations executed in low-resolution space.
One might question whether UHDPromer is also effective in general image restoration and enhancement tasks.
To answer this, we have conducted comparative analyses of our UHDPromer against standard image restoration and enhancement methods on various general image restoration and enhancement benchmarks, including low-light image enhancement on LOL~\citep{retinexnet_wei_bmvc18}, image dehazing on SOTS-ITS~\citep{RESIDE_dehazingbenchmarking_tip2019}, and image deblurring on GoPro~\citep{gopro2017}.
\begin{table*}[t]
% \vspace{-3mm}
\setlength{\tabcolsep}{23.75pt}
\caption{\textbf{Low-light image enhancement on LOL}~\citep{retinexnet_wei_bmvc18}.
Although our \textbf{UHDPromer} is designed for UHD images, where most operations are performed in the low-resolution space, our UHDPromer still achieves the best performance on general image enhancement benchmarks~\citep{retinexnet_wei_bmvc18}.
The term `UHDFour-2/8' refers to a variant of UHDFour with a shuffle-down factor of 2/8.
Conversely, `UHDFour-8' denotes the UHDFour version utilizing a shuffle-down factor of 8, specifically tailored for enhancing UHD low-light images.
}
\label{tab:Low-light image enhancement-lol-lolv2.} 
\begin{tabular}{l|c|ccccccc}
\Xhline{1.5pt}
\textbf{Method}&\textbf{Venue}&~~\textbf{PSNR}~$\uparrow$ & \textbf{SSIM}~$\uparrow$& \textbf{Parameters}~$\downarrow$
\\
\Xhline{1pt}
Retinex-Net~\citep{retinexnet_wei_bmvc18}&BMVC'18&16.77 &0.54  &0.45M \\
Zero-DCE~\citep{zerodce_lowlight_guo}&CVPR'20&16.79&0.67   & 79.416K\\
AGLLNet~\citep{AGLLNet}&IJCV'21&17.52& 0.77  &3.438K\\
Zhao et al.~\citep{zhao_lie}&ICCV'21&21.67 & 0.87  & 11.560M \\
RUAS~\citep{RUAS_liu_cvpr21}&CVPR'21&16.44 & 0.70   &3K\\
SCI~\citep{ma2022toward}&CVPR'22&14.78 & 0.62  &0.258K \\
URetinex-Net~\citep{Wu_2022_CVPR}&CVPR'22  &19.84&0.87 &1.5M\\
LLFlow~\citep{wang2021llflow}&AAAI'22& 28.99 & 0.94 & 17.4M \\
LLFormer~\citep{LLformer}&AAAI'23& 27.22 &0.90 &13.2M \\
UHDFour-2~\citep{Li2023ICLR_uhdfour}&ICLR'23& 23.09 &0.87   &28.518M\\
UHDFour-8~\citep{Li2023ICLR_uhdfour}&ICLR'23& 26.65 & 0.87 & 17.537M \\
UHDformer~\citep{uhdformer}&AAAI'24& 25.36 & 0.91  &0.3393M   \\
\Xhline{1.5pt}
\textbf{UHDPromer}&-&26.64 &0.91  &0.7430M  \\
\Xhline{1.5pt}
\end{tabular}
% \vspace{-2mm}
\end{table*}
\begin{table*}[!t]
\setlength{\tabcolsep}{23.75pt}
\caption{\textbf{Comparisons on SOTS-Indoor~\citep{RESIDE_dehazingbenchmarking_tip2019}. }
Although our \textbf{UHDPromer} is designed for UHD images, where most operations are performed in low-resolution space, our method is comparable with general image dehazing methods in terms of SSIM but lower in terms of PSNR.
}
\label{tab:Comparisons on SOTS-Indoor.} 
\begin{tabular}{l|c|ccccccc}
\Xhline{1.5pt}
\textbf{Method}&\textbf{Venue}& \textbf{PSNR}~$\uparrow$ & \textbf{SSIM}~$\uparrow$& \textbf{Parameters}~$\downarrow$
\\
\Xhline{1pt}
GridNet~\citep{grid_dehaze_liu}&ICCV'19&33.8244&0.9924&  0.96M \\
MSBDN~\citep{msbdn_cvpr20_dong}&CVPR'20&34.4219&0.9899 &31.4M \\
UHD~\citep{Zheng_uhd_CVPR21}&CVPR'21&18.7488 & 0.8486  &34.5M\\
PFDN~\citep{pfdn_eccv20_dong}&ECCV'20&31.2308 & 0.9853&12M\\
% PSD~\citep{psd_Chen_2021_CVPR}&CVPR'21&15.2682 & 0.7975&\\
D4~\citep{d4_dehze}&CVPR'22&20.5449 & 0.9236  &22.9M\\
DeHamer~\citep{guo2022dehamer}&CVPR'22&38.2672 & 0.9936&132.4M  \\
DehazeFormer~\citep{DehazeFormer}&TIP'23&39.9100&0.9960&25.5M\\
UHDformer~\citep{uhdformer}&AAAI'24& 35.7413  & 0.9920  &0.3393M   \\
\Xhline{1pt}
\textbf{UHDPromer}&-& 32.4778 & 0.9892 & 0.7430M\\
\Xhline{1.5pt}
\end{tabular}
\end{table*}
\begin{table*}[!t]
\setlength{\tabcolsep}{23.75pt}
\caption{\textbf{Comparisons on GoPro}~\citep{gopro2017}. 
Our UHDPromer is less effective in handling general image deblurring benchmarks, such as GoPro~\citep{gopro2017}.
Note that the PSNR and SSIM may be different from existing papers, as we use the commonly-used IQA PyTorch Toolbox (https://github.com/chaofengc/IQA-PyTorch) to recompute the results for all benchmarks in this paper.
}
\label{tab:Comparisons on GoPro.} 
\begin{tabular}{l|c|ccccccc}
\Xhline{1.5pt}
\textbf{Method}& \textbf{Venue} &  \textbf{PSNR}~$\uparrow$  &  \textbf{SSIM}~$\uparrow$ & \textbf{Parameters}~$\downarrow$ 
\\
\Xhline{1pt}
SRN~\citep{tao2018scale}&CVPR'18&31.751 & 0.9144 &6.8M  \\
DMPHN~\citep{dmphn2019}&CVPR'19&31.734&0.9130&21.7M  \\
MIMO-Unet++~\citep{cho2021rethinking_mimo}&ICCV'21&33.277 & 0.9348& 16.1M \\
MPRNet~\citep{Zamir_2021_CVPR_mprnet}&CVPR'21&34.195 & 0.9454& 20.1M  \\
Restormer~\citep{Zamir2021Restormer}&CVPR'22&34.483 & 0.9488 & 26.1M\\
Uformer~\citep{wang2021uformer}&CVPR'22&34.668 & 0.9502& 20.6M \\
Stripformer~\citep{Tsai2022Stripformer}&ECCV'22&34.678&0.9506&19.7M\\
FFTformer~\citep{Kong_2023_CVPR_fftformer}&CVPR'23&35.798&0.9594&16.6M\\
UHDformer~\citep{uhdformer}&AAAI'24&29.407 & 0.8660 &0.3393M\\
\Xhline{1pt}
\textbf{UHDPromer}&-&29.487 &0.8694  &0.7430M  \\
\Xhline{1.5pt}
\end{tabular}
% \vspace{-4mm}
\end{table*}
\subsubsection{Results on Low-light Image Enhancement}

Tab.~\ref{tab:Low-light image enhancement-lol-lolv2.} presents the comparative results of low-light image enhancement using the LOL dataset as referenced in \citep{retinexnet_wei_bmvc18}.
Despite our UHDPromer being tailored for Ultra-High Definition (UHD) images, primarily operating in low-resolution space, it still outperforms other methods which are specifically designed for general image enhancement~\citep{retinexnet_wei_bmvc18,AGLLNet,Wu_2022_CVPR}, as demonstrated on the widely recognized LOL benchmarks \citep{retinexnet_wei_bmvc18}.
% %
Our UHDPromer demonstrates competitive performance while maintaining superior computational efficiency compared to existing methods. 
%
% However, our method exhibits certain limitations. 
%
Our UHDPromer also slightly outperforms compared to UHDformer~\citep{uhdformer} according to PSNR metrics but underperforms specialized low-light enhancement methods designed for conventional image resolutions, e.g., UHDPromer achieves lower performance than strong baselines such as LLFlow~\citep{wang2021llflow} that contains 17.4M parameters, but it requires only 0.743M parameters-approximately 23.7 times fewer than LLFlow. Similarly, while achieving comparable SSIM performance to LLFormer~\citep{LLformer}, our method reduces 94.4\% of its parameters. Notably, LLFormer cannot directly process UHD images without additional preprocessing, whereas our UHDPromer natively handles UHD inputs with minimal computational overhead.
This performance gap suggests that while our approach excels in computational efficiency and UHD processing capabilities, there remains room for improvement in matching the restoration quality of task-specific architectures on benchmarks with general image sizes.
% \subsection{Results on GoPro~\cite{retinexnet_wei_bmvc18} for Low-light Image Enhancement}
%

\subsubsection{Results on Image Dehazing}
Tab.~\ref{tab:Comparisons on SOTS-Indoor.} presents our comparative study on image dehazing, specifically on the SOTS-Indoor dataset~\citep{RESIDE_dehazingbenchmarking_tip2019}.
Despite being tailored for UHD images with a primary focus on low-resolution space operations, our UHDPromer shows inferior performance in comparison with general image dehazing methods, which is specifically designed for image dehazing. 
It is noteworthy that UHDPromer requires only $0.7430$M parameters, which is significantly less than other state-of-the-art methods.
On the restoration performance for image dehazing, UHDPromer exhibits suboptimal performance on the SOTS-Indoor dataset~\citep{RESIDE_dehazingbenchmarking_tip2019} for image dehazing compared to UHDformer~\citep{uhdformer}. This performance pattern further confirms that our method's architectural design, while optimized for UHD processing efficiency, may not fully capture the complex degradation patterns present in general-resolution images across diverse restoration tasks.
\subsubsection{Results on Dynamic Image Deblurring}
% %
While our UHDPromer shows proficiency in handling Ultra-High Definition (UHD) images, it does have its limitations. Notably, UHDPromer struggles with general image deblurring tasks, such as those involving the GoPro dataset~\citep{gopro2017}, as detailed in Tab.~\ref{tab:Comparisons on GoPro.}. 
This limitation can be attributed to a couple of factors:
Firstly, general image deblurring often demands large model sizes to effectively manage complex blur. However, the UHDPromer is designed with a relatively modest parameter count of only 0.7430M, which may not be sufficient to adequately process intricate motion blur. 
Secondly, effective deblurring typically requires models to handle numerous operations on larger spatial features to achieve better results. Conversely, UHDPromer is tailored for UHD images, predominantly operating on 8$\times$ downsampling features. This design choice significantly impacts its deblurring performance in more general contexts.
Moreover, our UHDPromer slightly outperforms UHDformer~\citep{uhdformer} on the dynamic image deblurring benchmark~\citep{gong2017motion}. 
These comparisons further highlight the limitation of our UHDPromer, that is not suitable for handling general image sizes.
Our UHDPromer only works well on UHD image sizes, which limits its further application to more scenarios.
%%%%%%%%%%%%%%%%%%%%%
% \vspace{-8mm}
\section{Conclusion}
We have proposed a simple yet effective UHDPromer, which integrates the developed neural discrimination priors implying between high-resolution features and low-resolution ones into Transformer designs, for UHD image restoration and enhancement.
To better learn low-resolution features, we propose a sample and effective Neural Discrimination-Prompted Transformer Block that is guided by the developed NDP, containing Neural Discrimination-Prompted Attention (NDPA) and Neural Discrimination-Prompted Network (NDPN).
The NDPA effectively integrates the NDP into attention to perceive the useful discrimination information within global perspectives.
The NDPN is used to explore the continuous gating mechanism guided by the NDP to allow more beneficial information to be passed.
To better reconstruct the results, we have suggested a super-resolution-guided reconstruction approach.
Extensive experiments have demonstrated that our UHDPromer outperforms state-of-the-art methods under different training settings on $3$ UHD image restoration and enhancement tasks, including UHD low-light image enhancement, UHD dehazing, and UHD deblurring.
Additional experiments also show that our proposed UHDPromer is not good at handling general image size compared with both general image restoration methods.
This weakness limits our method for simultaneously handling general image size and UHD image size, thus limiting its broader applications.

In the future, we will focus on the potential works of how to keep the restoration performance for general image size while can handle UHD image size.
\\
\\
\noindent\textbf{Data Availability Statement.} 
All the data used in this manuscript is publicly available.
The source codes and pre-trained models will be also made available at \url{https://github.com/supersupercong/uhdpromer}.

{
% \small
%The spbasic bibliography style is designed to be used with natbib. So load natbib and then use \citept for Author (year) citations and \citepp for (Author, year) citations.
\bibliographystyle{spbasic}
\bibliography{egbib}
}

%%%%%%%%%%%%%%%%%%%%%%%%%%%%%%%%%%%%%%%%%%%%%%%%%%%%%%%%%%%%%
% To-Do
% 1 change cite to citep citet
% 2 table mutil-row fix

% Related info
% 1 http://www.uco.es/~in1majim/calls/ijcv3dhumans.html

% Add related work
% 1 3d model based 
%     hmr? video-based smpl
%     VIBE: Video Inference for Human Body Pose and Shape Estimation
% 2 2d pose to 3d pose
%     3D human pose estimation in video with temporal convolutions and semi-supervised training

% Add experiments

\end{document}